%% file: ICML_main_draft.tex
\documentclass{article}
\usepackage{multirow}

\usepackage{times}
\usepackage{url}            %
\usepackage{booktabs}       %
\usepackage{amsfonts}       %
\usepackage{nicefrac}       %
\usepackage{microtype}      %
\usepackage{mkolar_definitions}

\usepackage{xspace}
\usepackage{multirow}
\usepackage{amsmath}
\usepackage{algorithm}
\usepackage{algorithmic}
\usepackage{color}
\usepackage{authblk}
\usepackage{color, colortbl}
\usepackage{enumitem}
\usepackage{comment}
\usepackage{bm}
\usepackage{subcaption}

\input{notation}

\usepackage{microtype}
\usepackage{graphicx}
\usepackage{booktabs} %

\usepackage{hyperref}

\usepackage[accepted]{icml2024}

\usepackage{amsmath}
\usepackage{amssymb}
\usepackage{mathtools}
\usepackage{amsthm}

\usepackage[textsize=tiny]{todonotes}

\usepackage{subfiles}
\icmltitlerunning{Feedback Efficient Online Fine-Tuning of Diffusion Models}

\begin{document}

\twocolumn[
\icmltitle{Feedback Efficient Online Fine-Tuning of Diffusion Models}

\icmlsetsymbol{equal}{*}

\begin{icmlauthorlist}

\icmlauthor{Masatoshi Uehara}{equal,gen}
\icmlauthor{Yulai Zhao}{equal,pre}
\icmlauthor{Kevin Black}{ber}
\icmlauthor{Ehsan Hajiramezanali}{gen}
\icmlauthor{Gabriele Scalia}{gen}
\icmlauthor{Nathaniel Lee Diamant}{gen}
\icmlauthor{Alex M Tseng}{gen}
\icmlauthor{Sergey Levine$^{\dagger}$}{ber}
\icmlauthor{Tommaso Biancalani$^{\dagger}$}{gen}
\end{icmlauthorlist}

\icmlaffiliation{gen}{Genentech}
\icmlaffiliation{pre}{Princeton University }
\icmlaffiliation{ber}{University of California, Berkeley}

\icmlcorrespondingauthor{Sergey Levine}{sergey.levine@berkeley.edu}
\icmlcorrespondingauthor{Tommaso Biancalani }{biancalt@gene.com}

\icmlkeywords{Machine Learning, ICML}

\vskip 0.3in
]

\printAffiliationsAndNotice{\icmlEqualContribution} %

\begin{abstract}
Diffusion models excel at modeling complex data distributions, including those of images, proteins, and small molecules. However, in many cases, our goal is to model parts of the distribution that maximize certain properties: for example, we may want to generate images with high aesthetic quality, or molecules with high bioactivity. It is natural to frame this as a reinforcement learning (RL) problem, in which the objective is to fine-tune a diffusion model to maximize a reward function that corresponds to some property. Even with access to online queries of the ground-truth reward function, efficiently discovering high-reward samples can be challenging: they might have a low probability in the initial distribution, and there might be many infeasible samples that do not even have a well-defined reward (e.g., unnatural images or physically impossible molecules). In this work, we propose a novel reinforcement learning procedure that efficiently explores on the manifold of feasible samples. We present a theoretical analysis providing a regret guarantee, as well as empirical validation across three domains: images, biological sequences, and molecules. The code is available at \href{https://github.com/zhaoyl18/SEIKO}{https://github.com/zhaoyl18/SEIKO}. 
\end{abstract}

\section{Introduction}

\input{main_introduction}

\section{Related Works}
\input{main_related_works}

\section{Preliminaries}

\input{main_preliminary}

\section{Algorithm}

\input{main_proposed_algorithm}

\section{Regret Guarantees}

\input{main_regret}

\input{main_extensions}
\section{Experiments}

\input{main_experiments}

\input{main_future}

\bibliographystyle{chicago}
\bibliography{rl}

\appendix 

\newpage 

\onecolumn

\input{main_proof}

\end{document}

%% file: notation.tex
\newcommand{\alg}{\textbf{SEIKO}}
\newcommand{\pre}{\mathrm{pre}}
\newcommand{\KL}{\mathrm{KL}}
\newcommand{\condition}{\mathrm{con}}

\newcommand{\newedit}{\color{black}}

\definecolor{Gray}{gray}{0.9}

%% file: main_introduction.tex
\begin{figure}[!t]
    \definecolor{mypurple}{HTML}{FF41A0}
    \definecolor{mygreen}{HTML}{1CAF00}
    \definecolor{bananamania}{rgb}{0.98, 0.91, 0.71}
    \definecolor{bananayellow}{rgb}{1.0, 0.88, 0.21}
    \centering
    \includegraphics[width=1.0\linewidth]{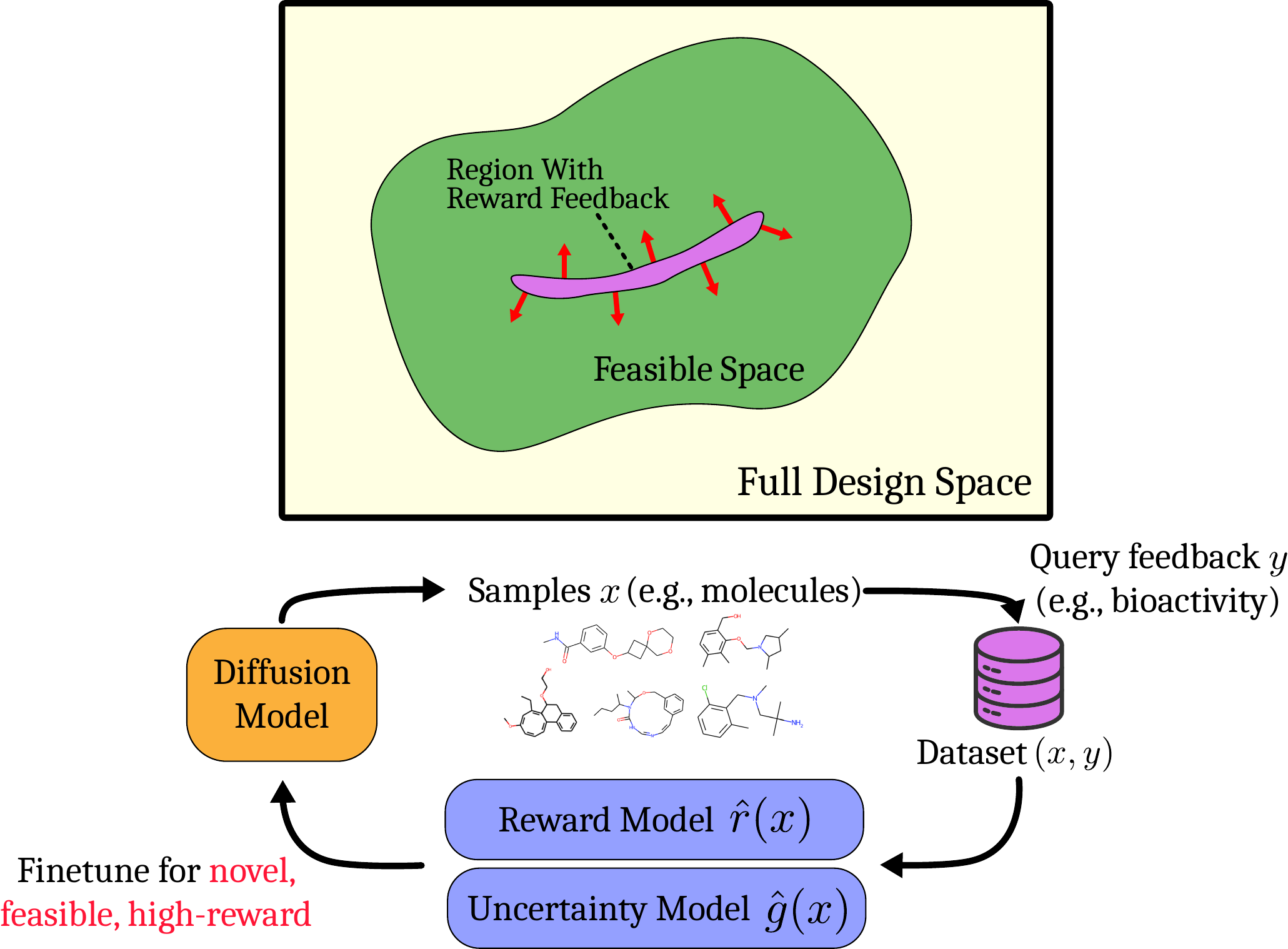}
    \caption{ We consider a scenario where we have a pre-trained diffusion model capturing a feasible region embedded as an intricate manifold (\textcolor{mygreen}{green}) in a high-dimensional full design space (\textcolor{bananayellow}{yellow}). We aim to fine-tune the model to get feasible samples with high rewards and novelty. The \textcolor{mypurple}{purple} area corresponds to the region where we have reward feedback, which we aim to expand with efficient online queries to the reward function.
    }
    \label{fig:online}
\end{figure}

Diffusion models belong to the family of deep generative models that reverse a diffusion process to generate data from noise. These diffusion models are highly adept at capturing complex spaces, such as the manifold of natural images, biological structures (e.g., proteins and DNA sequences), or chemicals (e.g., small molecules) \citep{watson2023novo,jing2022torsional,alamdari2023protein,zhang2024pre,igashov2022equivariant}. However, in many cases, our goal is to focus on parts of the distribution that maximize certain properties.
For example, in chemistry, one might train a diffusion model on a wide range of possible molecules, but applications of such models in drug discovery might need to steer the generation toward samples with high bioactivity. Similarly, in image generation, diffusion models are trained on large image datasets scraped from the internet, but practical applications often desire images with high aesthetic quality. This challenge be framed as a reinforcement learning (RL) problem, where the objective is to fine-tune a diffusion model to maximize a reward function that corresponds to the desirable property.

{
One significant obstacle in many applications is the cost of acquiring feedback from the ground-truth reward function. For instance, in biology or chemistry, evaluating ground truth rewards requires time-consuming wet lab experiments. In image generation, attributes such as aesthetic quality are subjective and require human judgment to establish ground truth scores. While several recent works have proposed methods for RL-based fine-tuning of diffusion models \citep{black2023training,fan2023dpok,clark2023directly,prabhudesai2023aligning,uehara2024finetuning}, none directly address the issue of feedback efficiency in an online setting. This motivates us to develop an online fine-tuning method that minimizes the number of reward queries. 

Achieving this goal demands efficient exploration. But, because we are dealing with high-dimensional spaces, this entails more than just exploring new regions. It also involves adhering to the structural constraints of the problem. For example, in biology, chemistry, and image generation, valid points such as physically feasible molecules or proteins typically sit on a lower-dimensional manifold (the ``feasible space'') embedded within a much larger full design space. Thus, an effective feedback-efficient fine-tuning method should explore the space without leaving this feasible area, as this would result in wasteful invalid queries.
}

Hence, we propose a feedback-efficient iterative fine-tuning approach for diffusion models that intelligently explores the feasible space, as illustrated in Figure~\ref{fig:online}.
Here, in each iteration, we acquire new samples from the current diffusion model, query the reward function, and integrate the result into the dataset. Then, using this augmented dataset with reward feedback, we train the reward function and its uncertainty model, which assigns higher values to areas not well-covered by the current dataset (i.e., more novel regions). We then update the diffusion model using the learned reward function, enhanced by its uncertainty model and a Kullback-Leibler (KL) divergence term relative to the current diffusion model, but without querying new feedback. This fine-tuned diffusion model then explores regions in the feasible space with high rewards and high novelty in the next iteration.

{ Our main contribution is to propose a provably feedback-efficient method for the RL-based online fine-tuning of diffusion models, compared to existing works focused on improving computational efficiency.}
{Our conceptual and technical innovation lies in introducing a novel online method tailored to diffusion models by {(a) interleaving reward learning and diffusion model updates 
and (b) integrating an uncertainty model and KL regularization to facilitate exploration while containing it to the feasible space.
Furthermore, we show the feedback efficiency of our method by providing a regret guarantee, and experimentally validate our proposed method on a range of domains, including images, protein sequences.

%% file: main_related_works.tex
Here, we offer a comprehensive review of related literature.

\paragraph{Fine-tuning diffusion models.} 

Many prior works have endeavored to fine-tune diffusion models by optimizing reward functions through supervised learning \citep{lee2023aligning,wu2023better}, policy gradient,  
\citep{black2023training,fan2023dpok}, or control-based (i.e., direct backpropagation) techniques \citep{clark2023directly,xu2023imagereward,prabhudesai2023aligning}.
Most importantly, these existing work assumes a static reward model; the rewards are either treated as ground truth, or if not, no allowance is made for online queries to ground-truth feedback.
In contrast, we consider an online setting in which additional queries to the ground-truth reward function are allowed; we then tackle the problems of exploration and feedback efficiency by explicitly including a reward modeling component that is interleaved with diffusion model updates.
Our experiments show improved feedback efficiency over prior methods across various domains, including images, biological sequences, and molecules, whereas prior work has only considered images.
Finally, our paper includes key theoretical results (Theorem~\ref{thm:key}) absent from prior work.
{\newedit
}

\paragraph{Online learning with generative models.} 

Feedback-efficient online learning has been discussed in several domains, such as NLP and CV \citep{brantley2021proceedings,zhang2022survey}. For example, in LLMs, fine-tuning methods that incorporate human feedback (RLHF), such as preferences, have gained widespread popularity \citep{touvron2023llama,ouyang2022training}. Based on that, \citet{dong2023raft} introduces a general online learning approach for generative models. However, their approach doesn't appear to be tailored specifically for diffusion models, which sets it apart from our work. Notably, their fine-tuning step relies on supervised fine-tuning, unlike our work. From a related perspective, recent studies \citep{yang2023using,wallace2023diffusion} have explored fine-tuning techniques of diffusion models using human feedback. However, their emphasis differs significantly in that they aim to leverage preference-based feedback by directly optimizing preferences without constructing a reward model following the idea of Direct Preference Optimization \citep{rafailov2024direct}. In contrast, our focus is on proposing a provably feedback-efficient method while still utilizing non-preference-based feedback.

\paragraph{Guidance.} 
\citet{dhariwal2021diffusion} introduced classifier-based guidance, an inference-time technique for steering diffusion samples towards a particular class. More generally, guidance uses an auxiliary differentiable objective (e.g., a neural network) to steer diffusion samples towards a desired property \citep{graikos2022diffusion,bansal2023universal}. However, we would expect this approach to work best when the guidance network is trained on high-quality data with reward feedback that comprehensively covers the sample space. Our emphasis lies in scenarios where we lack such high-quality data and must gather more in an online fashion. Indeed, we show in our experiments that our method outperforms a guidance baseline that merely uses the gradients of the reward model to steer the pre-trained diffusion model toward desirable samples.

\paragraph{Bandits and black-box optimization.} Adaptive data collection has been a topic of study in the bandit literature, and our algorithm draws inspiration from this body of work \citep{lattimore2020bandit}. However, in traditional literature, the action space is typically small. Our challenge lies in devising a practical algorithm capable of managing a complicated and large action space. We do this by integrating a pre-trained diffusion model. Although there exists literature on bandits with continuous actions using classical nonparametric methods \citep{krishnamurthy2020contextual,frazier2018tutorial,balandat2020botorch}, these approaches struggle to effectively capture the complexities of image, biological, or chemical spaces. Some progress in this regard has been made through the use of variational autoencoders (VAEs)  \citep{kingma2013auto} as shown in \citet{notin2021improving,gomez2018automatic}. However, our focus is on how to incorporate pre-trained diffusion models. 

Note that although there are several exceptions discussing black-box optimization with diffusion models \citep{wang2022diffusion,krishnamoorthy2023diffusion}, their fine-tuning methods rely on guidance and are more tailored to the offline setting. In contrast, we employ the RL-based fine-tuning method, and our setting is online. We will also compare our approach with guided methods in Section~\ref{sec:experiments}.

%% file: main_preliminary.tex
We consider an online bandit setting with a pre-trained diffusion model. Before explaining our setting, we provide an overview of our relation to bandits and diffusion models.

\paragraph{Bandits.}
Let's consider a design space denoted as $\Xcal$. Within $\Xcal$, each element $x$ is associated with a corresponding reward $r(x)$. Here, $r:\Xcal \to [0,1]$ represents an unknown reward function. A common primary objective is to strategically collect data and learn this reward function so that we can discover high-quality samples that have high $r(x)$. %
For example, in biology, our goal may be to discover protein sequences with high bioactivity.

As we lack knowledge of the reward function $r(\cdot)$, our task is to learn it by collecting feedback gradually. We focus on situations where we can access noisy feedback $y = r(x) + \epsilon$ for a given sample $x$. In real-world contexts, such as chemistry or biology, this feedback frequently comprises data obtained from wet lab experiments. Given the expense associated with obtaining feedback, our objective is to efficiently identify high-quality samples while minimizing the number of feedback queries.  This problem setting is commonly referred to as a bandit or black-box optimization \citep{lattimore2020bandit}.

\paragraph{Diffusion Models.}  
{\newedit A diffusion model is described by the following equation: 
\begin{align}
\label{eq:SDE}
    dx_t = f(t,x_t)dt + \sigma(t) d w_t,\, x_0\sim \nu, 
\end{align}
where $f:[0,T]\times \Xcal \to \Xcal^d$ is a drift coefficient, $\sigma:[0,T]\to \RR_{>0}$ is a diffusion coefficient associated with a $d$-dimensional Brownian motion $w_t$, and $\nu \in \Delta(\Xcal)$ is an initial distribution. It's worth noting that some papers use the reverse notation. When training diffusion models, the goal is to learn $f(t,x_t)$ from the data so that the generated distribution following the SDE aligns with the data distribution through score matching \citep{song2020denoising,ho2020denoising} or flow matching \citep{lipman2022flow,shi2023diffusion,tong2023improving,somnath2023aligned,albergo2023stochastic,liu20232,liu2022let}. 

In our study, we explore a scenario where we have a pre-trained diffusion model described by the following SDE:
\begin{align}
\label{eq:SDE_pretrained}
    dx_t = f^{\pre}(t,x_t)dt + \sigma(t) d w_t,\, x_0\sim \nu^{\pre} , 
\end{align}
where $f^{\pre}$ is a drift coefficient and $\nu^{\pre}$ is an initial distribution for the pre-trained model. We denote the generated distribution (i.e., marginal distribution at $T$ following \eqref{eq:SDE_pretrained}) by $p^{\pre}$.

\paragraph{Notation.}  We often use $[K]$ to denote $[1,\cdots, K]$. For $q,p\in \Delta(\Xcal)$, the notation $q \propto p $ means $q$ is equal to $p$ up to normalizing constants. We denote the KL divergence between $p,q\in \Delta(\Xcal)$ by $\KL(p\|q)$. We often consider a measure $\PP$ induced by an SDE on $\Ccal:=C([0,T],\Xcal)$ where $C([0,T],\Xcal)$ is the whole set of continuous functions on mapping from $[0,T]$ to $\Xcal$. When this SDE is associated with a drift term $ f$ and an initial distribution $\nu$ as in \eqref{eq:SDE}, we denote the measure by $\PP^{u,\nu}$. The notation $\EE_{\PP^{u,\nu}}[f(x_{0:T})]$ means that the expectation is taken for $f(\cdot)$ with respect to $\PP^{u,\nu}$. We denote the marginal distribution of $\PP^{u,\nu}$ over $\Xcal$ at time $t$ by $\PP^{u,\nu}_t$.  We also denote the distribution of the process conditioned on an initial and terminal point $x_0,x_T$ by $\PP_{\cdot|0,T}(\cdot |x_0, x_T)$ (we similarly define $\PP_{\cdot|T}(\cdot |x_T)$). With a slight abuse of notation, we exchangeably use distributions and densities.

In this work, we defer all proofs to Appendix~\ref{sec:proof}. 
}

\section{Problem Statement: Efficient RL Fine-Tuning of Diffusion Models}\label{sec:problem}

In this section, we state our problem setting: online bandit fine-tuning of pre-trained diffusion models. 

In contrast to standard bandit settings, which typically operate on a small action space $\Xcal$, our primary focus is on addressing the complexities arising from an exceedingly vast design space $\Xcal$. For instance, when our objective involves generating molecules with graph representation, the cardinality of $|\Xcal|$ is huge. 
While this unprocessed design space exhibits immense scale, the actual feasible and meaningful design space often resides within an intricate but potentially low-dimensional manifold embedded in $\Xcal$. In biology, this space is frequently referred to as the biological space. Similarly, in chemistry, it is commonly referred to as the chemical space. Notably, recent advances have introduced promising diffusion models aimed at capturing the intricacies of these biological or chemical spaces \cite{vignac2023digress,tseng2023graphguide,pmlr-v202-avdeyev23a}. We denote such a feasible space by $\Xcal_{\pre}$, as in Table~\ref{tab:problem}.

\begin{table}[!t]
    \centering
        \caption{Examples of original space and feasible space. }
    {
    \begin{tabular}{ccc}
      \toprule
         & Original space ($\Xcal$) & Feasible space ($\Xcal_{\pre}$) \\ \hline
 Images &  3-dimensional tensors & Natural images  \\ 
    Bio     & Sequence ($|20|^B,|4|^B$)  & Natural proteins/DNA \\ 
 Chem  & Graphs  &  Natural molecules  \\ 
                 \bottomrule
    \end{tabular}
}
    \label{tab:problem}
\end{table}

{ Now, let's formalize our problem. We consider feedback-efficient online fine-tuning of diffusion models. Specifically, we work on a bandit setting where we do not have any data with feedback initially, but we have a pre-trained diffusion model trained on a distribution $p^{\pre}(\cdot)$, whose support is given by $\Xcal_{\pre}$.  We aim to fine-tune this diffusion model to produce a new model $p \in \Delta(\Xcal)$ that maximizes $\EE_{x \sim p(x)}[r(x)]$, where $r(\cdot)$ is $0$ outside the support of $\Xcal_{\pre}$ ($\Xcal_{\pre}:=\{x\in \Xcal:p^{\pre}(x)>0 \}$) because a data not in $\Xcal_{\pre}$ is invalid. By leveraging the fine-tuned model, we want to efficiently explore the vast design space $\Xcal$ (i.e., minimizing the number of queries to the true reward function) while avoiding generation of invalid instances lying outside of $\Xcal_{\pre}$.
}

%% file: main_proposed_algorithm.tex
\begin{algorithm}[!t]
\caption{\alg\,(Optimi\textbf{S}tic fin\textbf{E}-tuning of d\textbf{I}ffusion with \textbf{K}L c\textbf{O}nstraint)}\label{alg:main}
\begin{algorithmic}[1]
{\newedit 
  \STATE {\bf Require}: Parameter $\alpha, \{\beta_i\} \in \RR^{+}$, a pre-trained diffusion model described by $f^{\pre}:[0,T]\times \Xcal \to \Xcal$, an initial distribution $\nu^{\pre}:\Xcal \to \Delta(\Xcal)$
  \STATE   Initialize $f^{(0)} = f^{\pre}, \nu^{(0)} = \nu^{\pre}$
  \FOR{$i$ in $[1,\cdots,K]$}
      \STATE Generate a new sample $x^{(i)} \sim p^{(i-1)}(x)$ following 
      \begin{align}\label{eq:sde_i}
          dx_t = f^{(i-1)}(t,x_t)dt + \sigma(t) dw_t,\,\, x_0 \sim \nu^{(i-1)}, 
      \end{align}
      from $0$ to $T$, and get a feedback $y^{(i)}=r(x^{(i)}) + \epsilon$. (Note $p^{(0)}=p^{\pre}$.) \label{lst:step1} 
      \STATE Construct a new dataset: $\Dcal^{(i)} = \Dcal^{(i-1)}\cup  (x^{(i)},y^{(i)})$ \label{lst:step4}
      \STATE Train a reward model $\hat r^{(i)}(x)$ and uncertainty oracle $\hat g^{(i)}(x)$ using a dataset $\Dcal^{(i)}$  \label{lst:step2}   
      \STATE  \label{lst:step3}
      
       Update a diffusion model  by solving the control problem: 
       {\tiny
      \begin{align}\label{eq:key}
        & f^{(i)}, \nu^{(i)}=    \argmax_{f:[0,T]\times \Xcal \to \Xcal,\nu \in \Delta(\Xcal) }\underbrace{\EE_{\PP^{f,\nu}}[(\hat r^{(i)}+\hat g^{(i)})(x_T)]}_{\text{(B) Optimistic\,reward}}   \\ 
         & \textstyle  - \alpha \underbrace{\EE_{\PP^{f,\nu}}\left [ \log \left (\frac{\nu(x_0)}{\nu^{\pre}(x_0)} \right) + \int_{t=0}^T\frac{\|g^{(0)}(t,x_t)\|^2 }{2\sigma^2(t)}dt  \right ]}_{\text{(A1) KL\,Regularization\,relative\,to\,a diffusion at iteration $0$ } }   \nonumber \\
         &- \textstyle \beta_{i-1}   \underbrace{  \EE_{\PP^{f,\nu}}\left [\log\left(\frac{\nu(x_0)}{\nu^{(i-1)}(x_0)}\right)+ \int_{t=0}^T\frac{\|g^{(i-1)}(t,x_t)\|^2 }{2\sigma^2(t)}dt  \right ] }_{\text{(A2) KL\,Regularization\,relative\,to\,a diffusion at iteration $i-1$}  }  \nonumber 
      \end{align}
      } 
      where $g^{(i-1)}:=  f^{(i-1)}- f$, $g^{(0)}:=  f^{(0)}- f  $ and the expectation $\EE_{\PP^{f,\nu}} [\cdot]$ is taken with respect to the distribution induced by the SDE associated with a drift $f$ and an initial distribution $\nu$ in \eqref{eq:SDE}. Refer to Appendix~\ref{sec:neuralSDE} regarding algorithms to solve \eqref{eq:key}. 
  \ENDFOR 
  \STATE {\bf Output}:  $p^{(1)},\cdots,p^{(K)}$.  
  } 
\end{algorithmic}
\end{algorithm}

In this section, we present our novel framework \alg\,for fine-tuning diffusion models. { The core of our method consists of two key components: (a) effectively preserving the information from the pre-trained diffusion model through KL regularization, enabling exploration within the feasible space $\Xcal_{\pre}$, (b) introducing an optimistic bonus term to facilitate the exploration of novel regions of $\Xcal_{\pre}$. } 

Our algorithm follows an iterative approach. Each iteration comprises three key steps: (i) feedback collection phase, (ii) reward model update with feedback, and (iii) diffusion model update. {Here, we decouple the feedback collection phase (i) and diffusion model update (iii) so that we do not query new feedback to the true reward when updating the diffusion model in step (iii).}
Algorithm~\ref{alg:main} provides a comprehensive outline of \alg\ (Optimistic Finetuning of Diffusion models with KL constraint). We will elaborate on each component in the subsequent sections.

\subsection{Data Collection Phase  (Line~\ref{lst:step1})}

We consider an iterative procedure. Hence, at this iteration $i \in [K]$, we have a fine-tuned diffusion model (when $i=1$, this is just a pre-trained diffusion model), that is designed to explore designs with high rewards and high novelty in Subsection~\ref{subsec:diffusion}.  

{\newedit Using this diffusion model, we obtain new samples in each iteration and query feedback for these samples. To elaborate, in each iteration, we maintain a drift term $f^{(i-1)}$ and an initial distribution $\nu^{(i-1)}$. Then, following the SDE associated with $f^{(i-1)}$ and $\nu^{(i-1)}$ in \eqref{eq:sde_i}, we construct a data-collection distribution 
$p^{(i-1)} \in \Delta(\Xcal)$ (i.e. a marginal distribution at $T$).  After getting a sample $x^{(i)}$, we obtain its corresponding feedback $y^{(i)}$. Then, we aggregate this new pair $\{x^{(i)},y^{(i)}\}$ into the existing dataset $\Dcal^{(i-1)}$ in Line~\ref{lst:step4}. }

\begin{remark}[Batch online setting] \label{rem:batch}
To simplify the notation, we present an algorithm for the scenario where one sample is generated during each online epoch. In practice, multiple samples may be collected. In such cases, our algorithm remains applicable, and the theoretical guarantee holds with minor adjustments \citep{gao2019batched}. 
\end{remark}

\subsection{Reward Model Update (Line~\ref{lst:step2}) }

We learn a reward model $\hat r^{(i)}:\Xcal \to \RR $ from the dataset $\Dcal^{(i)}$. To construct $\hat r$, a typical procedure is to solve the (regularized) empirical risk minimization (ERM) problem: 
\begin{align}\label{eq:ERM}
  \hat r^{(i)}(\cdot) =  \argmin_{r \in \Fcal}\sum_{(x,y)\sim \Dcal^{(i)} } \{r(x)- y\}^2 + \|r\|_{\Fcal},
\end{align}
where $\Fcal$ is a hypothesis class such that $\Fcal \subset [\Xcal \to \RR]$ and $\|\cdot\|_{\Fcal}$ is a certain norm to define a regularizer. 

We also train an uncertainty oracle, denoted as $\hat g^{(i)}:\Xcal \to [0,1]$, using the data $\Dcal^{(i)}$. The purpose of this uncertainty oracle $\hat g^{(i)}$ is to assign higher values when $\hat r^{(i)}$ exhibits greater uncertainty. We leverage this uncertainty oracle $\hat g^{(i)}$ to facilitate exploration beyond the current dataset when we update the diffusion model, as we will see in Section~\ref{subsec:diffusion}. This can be formally expressed as follows.

 \begin{definition}[Uncertainty oracle]\label{assum:calibrated}
 With probability $1-\delta$, 
\begin{align*} \textstyle
  x \in \Xcal_\pre;\, |\hat r^{(i)}(x) - r(x)|\leq  \hat g^{(i)}(x). 
\end{align*}
Note that the above only needs to hold within $\Xcal_\pre$. 
 \end{definition}

Provided below are some illustrative examples of such $\hat g^{(i)}$.

\begin{example}[Linear models]\label{exm:linear}
When we use a linear model $\{ x \mapsto \theta^{\top}\phi(x)\}$ with feature vector $\phi:\Xcal \to \RR^d$ for $\Fcal$, we use $(\hat r^{(i)},\hat g^{(i)})$ as follows: 
\begin{align*}\textstyle 
    \hat r^{(i)}(\cdot)  & =\phi^{\top}(\cdot)(\Sigma_i+\lambda I)^{-1} \sum_{j=1}^i \phi(x^{(j)})y^{(j)}, \\ 
    \hat g^{(i)}(\cdot) &= C_1(\delta) \min\left (1,  \sqrt{\phi(\cdot)^{\top}(\Sigma_i+\lambda I)^{-1} \phi(\cdot)}\right), 
\end{align*}
where $\Sigma_i = \sum_{j=1}^i \phi(x^{(j)})\phi(x^{(j)})^{\top}$ , $C_1(\delta) , \lambda \in \RR^{+}$. This satisfies Definition~\ref{assum:calibrated} with a proper choice of $C_1(\delta) $ as explained in \citet{agarwal2019reinforcement} and Appendix~\ref{ape:prerlim}. This can also be extended to cases when using RKHS (a.k.a. Gaussian processes) as in \citet{srinivas2009gaussian,garnelo2018neural,valko2013finite,chang2021mitigating}. 

A practical question is how to choose a feature vector $\phi$. In practice, we recommend using the last layer of neural networks as $\phi$ \citep{zhang2022making,qiu2022contrastive}.  
\end{example}

\begin{example}[Neural networks] \label{exm:nn}
When we use neural networks for $\Fcal$, a typical construction of $\hat g^{(i)}$ is based on statistical bootstrap \citep{efron1992bootstrap}. Many practical bootstrap methods with neural networks have been proposed in \citet{osband2016deep,chua2018deep}, and its theory has been analyzed \citep{kveton2019garbage}. Specifically, in a typical procedure, given a dataset $\mathcal{D}$, we resample with a replacement for $M$ times and get $\mathcal{D}_1, \cdots, \mathcal{D}_M$. We train an individual model $\{\hat r^{(i,j)}\}_{j=1}^M$ with each bootstrapped dataset. Then, we set $\hat r^{(i)} = \max_{j\in [M]}  \hat r^{(i,j)}$ for $i \in [K]$. 
\end{example}

\subsection{Diffusion Model Update (Line~\ref{lst:step3}) } \label{subsec:diffusion}

In this stage, without querying new feedback, we update the diffusion models (i.e., drift coefficient $f^{(i-1)}$). Our objective is to fine-tune a diffusion model to generate higher-reward samples, exploring regions not covered in the current dataset while using pre-trained diffusion models to avoid deviations from the feasible space. 

{\newedit To achieve this goal, we first introduce an optimistic reward term (i.e., reward with uncertainty oracle) to sample high-reward designs while encouraging exploration. We also include a KL regularization term to prevent substantial divergence between the updated diffusion model and the current one at $i-1$. This regularization term also plays a role in preserving the information from the pre-trained diffusion model, keeping exploration constrained to the feasible manifold $\Xcal_{\pre}$. The parameters $\beta_i,\alpha$ govern the magnitude of this regularization term.

Formally, we frame this phase as a control problem in a special version of soft-entropy-regularized MDPs \citep{neu2017unified,geist2019theory}, as in Equation~\pref{eq:key}. In this objective function, we aim to optimize three terms: (B) optimistic reward at the terminal time $T$, (A1) KL term relative to a pre-trained diffusion model, and (A2) KL term relative to a diffusion model at $i-1$. Indeed, for terms (A1) and (A2), we use observations:  
\begin{align*}
    \mathrm{(A1)} &=\KL(\PP^{f,\nu}(\cdot) \| \PP^{f^{\pre},\nu^{\pre}} (\cdot)), \\  \mathrm{(A2)} &=\KL(\PP^{f,\nu}(\cdot)\| \PP^{f^{(i-1)},\nu^{(i-1)}}(\cdot) ). 
\end{align*}

Importantly, we can show that we are able to obtain a more explicit form for $p^{(i)}(\cdot)$, which is a distribution generated by the fine-tuned diffusion model with $f^{(i)},\nu^{(i)}$ after solving \eqref{eq:key}. Indeed, we design a loss function \eqref{eq:key} to obtain this form so that we can show our method is provably feedback efficient in \pref{sec:regret}. 

 \begin{theorem}[Explicit form of fine-tuned distributions]\label{thm:key}
The distribution $p^{(i)}$ satisfies
{\small 
\begin{align}\label{eq:implicit}
    p^{(i)} = \argmax_{p \in \Delta(\Xcal)}\underbrace{\EE_{x\sim p}[r(x)]}_{\mathrm{(B')}}- \alpha \underbrace{\KL(p\|p^{(0)})}_{\mathrm{(A1')}} - \beta \underbrace{\KL(p\|p^{(i-1)})}_{\mathrm{(A2')} }. 
\end{align}
} 
It is equivalent to 
\begin{align}\label{eq:explicit}
p^{(i)} \propto  \exp\left(\frac{\hat r^{(i)}(\cdot)+\hat g^{(i)}(\cdot)}{\alpha +\beta}\right)\{p^{(i-1)}(\cdot)\}^{\frac{\beta}{\alpha +\beta}}\{p^{\pre}(\cdot)\}^{\frac{\alpha}{\alpha +\beta}}. 
\end{align}
\end{theorem}

Equation~\eqref{eq:implicit} states that the $p^{(i)}$ maximizes the function, which comprises three terms: (B'), (A1'), and (A2'), which appear to be similar to (B), (A1), (A2) in \eqref{eq:key}. Indeed, the term (B') is essentially equivalent to (B) by regarding $p$ as the generated distribution at $T$ by $\PP^{f,\nu}$. As for terms (A1') and (A2'), the KL divergences are now defined on the distribution of $x_T$ (over $\Xcal$) unlike (A1), (A2) that are defined on trajectories on $x_{0:T}$ (over $\Ccal$). To bridge this gap, in the proof, we leverage key bridge preserving properties: $\mathrm{(A1')}-\mathrm{(A1)}=0$ and $\mathrm{(A2')}-\mathrm{(A2)}=0$. These properties stem from the fact that the posterior distribution over $\Ccal$ on $x_T$ is the same, leading to the conditional KL divergence over $x_{0:T}$ conditioning on $x_T$ being $0$.  

Equation~\eqref{eq:explicit} indicates that $p^{(i)}$ is expressed as the product of three terms: the optimistic reward term, the distribution by the current diffusion model $p^{(i-1)}$, and the pre-trained diffusion model. Through induction, it becomes evident that the support of $p^{(i)}$ always falls within that of the pre-trained model. Consequently, we can avoid invalid designs.

Some astute readers might wonder why we don't directly sample from $p^{(i)}$ in \eqref{eq:explicit}. This is because even if we know a drift term $f^{\pre}$, we lack access to the exact form of $p^{\pre}$ itself. In our algorithm, without directly trying to estimate $p^{\pre}$ and solving the sampling problem, we reduce the sampling problem to a control problem in Equation~\eqref{eq:key}. 
}

\paragraph{Algorithms to solve \eqref{eq:key}.}
 To solve \eqref{eq:key} practically, we can leverage a range of readily available control-based algorithms. Since $\hat r+\hat g$ is differentiable, we can approximate the expectation over trajectories, $\EE_{f,\nu}[\cdot]$, using any approximation method, like Euler-Maruyama, and parameterize $f$ and $\nu$ with neural networks. This enables direct optimization of $f$ and $\nu$ with stochastic gradient descent, as used in \citet{clark2023directly,prabhudesai2023aligning,uehara2024finetuning}. For details, refer to Appendix~\ref{sec:neuralSDE}. We can also use PPO for this purpose \citet{fan2023dpok,black2023training}.

%% file: main_regret.tex
\label{sec:regret}

{ %

 We will demonstrate the provable efficiency of our proposed algorithm in terms of the number of feedback iterations needed to discover a nearly optimal generative model. We quantify this efficiency using the concept of ``regret''. 

{\newedit To define this regret, first, we define the performance of a generative model $p \in \Delta(\Xcal)$ as the following value:
\begin{align}\label{eq:score}
    J_{\alpha}(p):= \underbrace{\EE_{x\sim p}[r(x)]}_{\mathrm{(B)}}- \alpha \underbrace{\KL(p\|p^{\pre}) }_{\mathrm{(A1')}}. 
\end{align}
The first term (B) represents the expected reward of a generative model $p$. The second term (A1') acts as a penalty when generative models generate samples that deviate from natural data (such as natural molecules or foldable proteins) in the feasible space, as we mention in \pref{sec:problem}. We typically assume that $\alpha$ is very small. This $\alpha$ controls the strength of the penalty term. However, even if $\alpha$ is very small, when $p$ is not covered by $p^{\pre}$, a generative model will incur a substantial penalty, which could be $\infty$ in extreme cases. 

Now, using this value, regret is defined as a quantity that measures the difference between the value of a diffusion model trained with our method and that of a generative model $\pi$ we aim to compete with (i.e., a high-quality one). While regret is a standard metric in online learning \citep{lattimore2020bandit}, we provide a novel analysis tailored specifically for diffusion models, building upon \pref{thm:key}. For example, while a value is typically defined as $\EE_{x\sim \pi}[r(x)]$ in the standard literature on online learning, we consider a new value tailored to our context with pre-trained diffusion models by adding a KL term.   
}

\begin{theorem}[Regret guarantee] \label{thm:regret}
With probability $1-\delta K$, by taking $\beta_i=\alpha(i-1)$, we have the following regret guarantee: $ \forall \pi\in \Delta(\Xcal);$
\begin{align*}
\underbrace{J_{\alpha}(\pi) - \frac{1}{K} \sum_{k=1}^K J_{\alpha}(p^{(i)})}_{\mathrm{Regret}} \leq %
 \underbrace{\frac{2}{K} \sum_{i=1}^K  \EE_{x \sim p^{(i)}}[\hat g^{(i)}(x) ]}_{ \mathrm{Statistical\,Error}}. 
\end{align*} 
\end{theorem}

This theorem indicates that the performance difference between any generative model $\pi \in \Delta(\Xcal)$ and our fine-tuned generative model is effectively controlled by the right-hand side. %
The right hand side corresponds to statistical error, which typically diminishes as we collect more data, often following rates on the order of $O(1/\sqrt{K})$ under common structural conditions. %
More specifically, the statistical error term, which depends on the complexity of the function class $\Fcal$ { because uncertainty oracles yield higher values as the size of the function class $\Fcal$ expands.} 
While the general form of this error term in \pref{thm:regret} might be hard to interpret, it simplifies significantly for many commonly used function classes. For example, if the true reward is in the set of linear models and we use linear models to represent $\hat{r}$, the statistical error term can be bounded as follows:

\begin{corollary}[Statistical error for linear model]\label{lem:linear}
When we use a linear model as in Example~\ref{exm:linear} such that $\forall x;\|\phi(x)\|_2\leq B$, suppose there exists a function that matches with $r:\Xcal \to \RR$ on the feasible space $\Xcal_{\pre}$ (i.e., realizability in $\Xcal_{\pre}$). By taking $\lambda=1$ and $\delta=1/H^2$, the expected statistical error term in \pref{thm:regret} is upper-bounded by $ \tilde O(d/\sqrt{K})$ where $\tilde O(\cdot)$ hides logarithmic dependence.  
\end{corollary}
Note when we use Gaussian processes and neural networks, we can still replace $d$ by effective dimension \citep{valko2013finite,zhou2020neural}.  
} 

{\newedit \begin{remark}[$\alpha$ needs to be larger than $0$] 
In Theorem~\ref{thm:regret}, it is important to set $\alpha>0$. When $\alpha=0$, the support of $p^{(i)}$ is not constrained to remain within $\Xcal_{\pre}$. However, $r(\cdot)$ takes $0$ outside of $\Xcal_{\pre}$ in our context (See Section~\ref{sec:problem}). 
\end{remark}
}

%% file: main_experiments.tex
\label{sec:experiments}

{
{ Our experiments aim to evaluate our proposed approach in three different domains. We aim to investigate the effectiveness of our approach over existing baselines and show the usefulness of exploration with KL regularization and optimistic exploration. } 

To start, we provide an overview of these baselines, detail the experimental setups, and specify the evaluation metrics used across all three domains. For comprehensive information on each experiment, including dataset and architecture details and additional results, refer to  Appendix~\ref{ape:experiments}. The code is available at \href{https://github.com/zhaoyl18/SEIKO}{https://github.com/zhaoyl18/SEIKO}

\vspace{-2mm}
{ \paragraph{Methods to compare.}  We compare our method with four baseline fine-tuning approaches. The first two are non-adaptive methods that optimize the reward without incorporating adaptive online data collection. The second two are na\"ive online fine-tuning methods.

\setlist{nolistsep}
\begin{itemize}
    \item \textbf{Non-adaptive \citep{clark2023directly,prabhudesai2023aligning} }: We gather $M$ samples (i.e., $x$) from the pre-trained diffusion model and acquire the corresponding feedback (i.e., $y$). Subsequently, after training a reward model $\hat r$ from this data with feedback, we fine-tune the diffusion model using this static $\hat r$ with direct back-propagation. 
    \item \textbf{Guidance (a.k.a. classifier guidance) \citep{dhariwal2021diffusion} }:
    We collect $M$ samples from the pre-trained diffusion model and receive their feedback $y$. We train a reward model on this data and use it to guide the sampling process toward high rewards.
 \item \textbf{Online (model-free) PPO:} We run KL-penalized RL finetuning with PPO \citep{schulman2017proximal} for $M$ reward queries. This is an improved version of both DPOK \citep{fan2023dpok} and DDPO \citep{black2023training} with a KL penalty applied to the rewards. For details, refer to Appendix~\ref{subsec:baselines}.
    \item \textbf{$\alg$}:
We collect $M_i$ data at every online epoch in $i \in [K]$ so that the total feedback budget (i.e., $\sum_{i=1}^K M_i$) is $M$ as in Remark~\ref{rem:batch}.  
    \begin{itemize}
     \item \textbf{Greedy (Baseline)}: Algorithm~\ref{alg:main} by setting $\beta =0$ (no KL term) and $\hat g$ to be $0$. 
     \item \textbf{Our proposal} with $\beta>0$ and $\hat g\neq 0$. 
     \begin{itemize}
        \item \textbf{UCB}: Algorithm~\ref{alg:main} with $\hat g$ as in  Example~\ref{exm:linear}.
        \item \textbf{Bootstrap}: Algorithm~\ref{alg:main} with  $\hat g$ as in  Example~\ref{exm:nn}.
     \end{itemize}
    \end{itemize}
\end{itemize}
} 

\vspace{-2mm}
\paragraph{Evaluation.} We ran each of the fine-tuning methods 10 times, except for image tasks. In each trial, we trained a generative model $p\in \Delta(\RR^d)$. For this $p$, we measured two metrics: the value, $J_{\alpha}(p)=\EE_{x\sim p}[r(x)]-\alpha \KL(p \|\rho_{\pre})$ in \eqref{eq:score} with a very small $\alpha$ (i.e., $\alpha=10^{-5}$), and the diversity among the generated samples, defined as $\EE_{x\sim p,z\sim p}[d(x,z)]$, where $d(x,z)$ quantifies the distance between $x$ and $z$. . Finally, we reported the mean values of (Value) and (Div) across trials. While our objective is not necessarily focused on just obtaining diverse samples, we included diversity metrics following common conventions in biological tasks.

\vspace{-2mm}
\paragraph{Setup.} {We construct a diffusion model by training it on a dataset that includes both high and low-reward samples, which we then employ as our pre-trained diffusion model. Next, we create an oracle $r$ by training it on an extensive dataset with feedback. In our experimental setup, we consider a scenario in which we lack knowledge of $r$. However, given $x$, we can get feedback in the form of $y = r(x)+\epsilon$. Our aim is to obtain a high-quality diffusion model that achieves high $\EE_{x \sim p}[r(x)]$ by fine-tuning using a fixed feedback budget $M$.
}

\subsection{Protein Sequences   }\label{subsec:biology}

\begin{table}[!t]
    \centering
      \caption{Results for fine-tuning diffusion models for protein sequences using the GFP to optimize fluorescence properties. \alg\, attains high rewards using a fixed budget of feedback.  }
    \begin{tabular}{c c  c } \toprule 
         & Value $\uparrow$   & Div $\uparrow$  \\  \midrule 
    \textbf{Non-adaptive} & $3.66 \pm 0.03$   &  $2.5$   \\
    \textbf{Guidance} &  $3.62 \pm 0.02$ &   $\mathbf{2.7}$  \\ \hline 
\textbf{Online PPO} & $3.63 \pm 0.04$   &  $2.5$\\ 
 \textbf{Greedy} &  $3.62 \pm 0.00$  & $2.4$ \\ \hline 
\rowcolor{Gray}    \textbf{UCB (Ours) } & $3.84\pm 0.00$  & $2.6$     \\
  \rowcolor{Gray}   \textbf{Bootstrap (Ours)}  &  $\mathbf{3.87\pm 0.01}$  & $2.2$ \\ \bottomrule
    \end{tabular}
    \label{tab:GFP}  
\end{table}

Our task involves obtaining a protein sequence with desirable properties. As a biological example, we use the GFP dataset \citep{sarkisyan2016local,trabucco2022design}. In the GFP task, $x$ represents green fluorescent protein sequences, each with a length of 237, and a true $r(x)$ denotes their fluorescence value. We leverage a transformer-based pre-trained diffusion model and oracle using this dataset. To measure distance, we adopt the Levenshtein distance, following \citet{ghari2023generative}. We set $M_i=500$ and $K=4$ ($M=2000$).

\paragraph{Results.} We report results in Table~\ref{tab:GFP}. First, it's evident that our online method outperforms non-adaptive baselines, underscoring the importance of adaptive data collection like our approach. Second, in comparison to online baselines such as \textbf{Online PPO} and \textbf{Greedy}, our approach consistently demonstrates superior performance. This highlights the effectiveness of our strategies, including (a) interleaving reward learning and diffusion model updates, (b) incorporating KL regularization and optimism for feedback-efficient fine-tuning. Third, among our methods,
\textbf{Bootstrap} appears to be the most effective.

\begin{remark}[Molecules]
We do an analogous experiment in Section~\ref{subsec:biology} to generate molecules with improved properties, specifically the Quantitative Estimate of Druglikeness (QED) score. For details, refer to Section~\pref{subsec:molecules}. 
\end{remark}

{

\subsection{Images}

We validate the effectiveness of \textbf{\alg} in the image domain by fine-tuning for aesthetic quality. We employ Stable Diffusion v1.5 as our pre-trained model~\citep{Rombach_2022_CVPR}.
As for $r(x)$, we use the LAION Aesthetics Predictor V2~\citep{schuhmann2022laion}, which is implemented as a linear MLP on the top of the OpenAI CLIP embeddings~\citep{radford2021clip}. This predictor is trained on a dataset comprising over $400$k aesthetic ratings from $1$ to $10$, and has been used in existing works \citep{black2023training}. In our experiment, we set $M_1 = 1024$, $M_2 = 2048$, $M_3 = 4096$, and $M_4 = 8192$. %

\begin{figure}[!t]
    \centering
    \includegraphics[width = \linewidth]{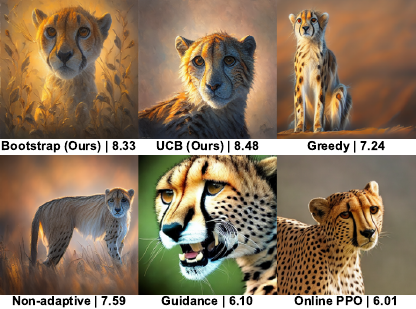}
    \caption{Images generated with aesthetic scores using the prompt ``cheetah." Our methods outperform the baselines in terms of higher aesthetic scores while using the same amount of feedback.}
    \label{fig:images}
\end{figure}

\begin{table}[!t]
    \centering
      \caption{Results for fine-tuning Stable Diffusion to optimize aesthetic scores. \alg\, attains high rewards within a fixed budget. }
    \begin{tabular}{p{0.40\linewidth}p{0.23\linewidth} }
    \toprule 
         & Value $\uparrow $   \\  \midrule 
    \textbf{Non-adaptive} & $7.22 \pm 0.18$    \\
    \textbf{Guidance} & $5.78 \pm 0.28$   \\ \hline 
\textbf{Online PPO} & $6.13 \pm 0.33$  \\ 
 \textbf{Greedy} &  $6.24 \pm 0.28$   \\ \hline 
   \rowcolor{Gray} \textbf{UCB (Ours) } & $\mathbf{8.17 \pm 0.33}$    \\
  \rowcolor{Gray}   \textbf{Bootstrap (Ours)}  &  $7.77\pm  0.12$  \\
    \bottomrule
    \end{tabular}
    \label{tab:fig_table}  
\end{table}

\vspace{-2mm}
\paragraph{Results.} We present the performances of \alg\, and baselines in Table~\ref{tab:fig_table}. Some examples of generated images are shown in Figure~\ref{fig:images}. 
It is evident that $\alg$ outperforms \textbf{Non-adaptive} and \textbf{Guidance} with better rewards, 
showcasing the necessity of adaptively collecting samples.
Moreover, \alg\  achieves better performances even compared to \textbf{Greedy} and \textbf{Online PPO}. This again demonstrates that incorporating KL regularization and optimism are useful for feedback-efficient fine-tuning. 

Finally, we present more qualitative results in Figure~\ref{fig:more-samples} to demonstrate the superiority of $\alg$ over all baselines. For more additional images, refer to Appendix~\ref{ape:experiments}. 
} 
}

%% file: main_future.tex
\section{Conclusion}

This study introduces a novel feedback-efficient method to fine-tune diffusion models. Our approach is able to explore a novel region with a high reward on the manifold, which represents the feasible space. Our theoretical analysis highlights the effectiveness of our approach in terms of feedback efficiency, offering a specific regret guarantee. 

As our future work, expanding on Section~\ref{subsec:biology}, we will explore the fine-tuning of recent diffusion models that are more customized for biological or chemical applications \citep{li2023latent,gruver2023protein,luo2022antigen}.

\section*{Impact Statement}

This paper presents work whose goal is to advance the field of Machine Learning. There are many potential societal consequences of our work, none of which we feel must be specifically highlighted here. 

%% file: main_proof.tex
\section{Implementation Details of Planning Algorithms}\label{sec:neuralSDE}

We seek to solve the following optimization problem:
\begin{align*}
        & f^{(i)}, \nu^{(i)}=    \argmax_{f:[0,T]\times \Xcal \to \Xcal,\nu \in \Delta(\Xcal) }\underbrace{\EE_{\PP^{f,\nu}}[(\hat r^{(i)}+\hat g^{(i)})(x_T)]}_{\text{(B) Optimistic\,reward}}   \\ 
         & \textstyle  - \alpha \underbrace{\EE_{\PP^{f,\nu}}\left [ \log \left (\frac{\nu(x_0)}{\nu^{\pre}(x_0)} \right) + \int_{t=0}^T\frac{\|g^{(0)}(t,x_t)\|^2 }{2\sigma^2(t)}dt  \right ]}_{\text{(A1) KL\,Regularization\,relative\,to\,a diffusion at iteration $0$ } }- \textstyle \beta_{i-1}   \underbrace{  \EE_{\PP^{f,\nu}}\left [\log\left(\frac{\nu(x_0)}{\nu^{(i-1)}(x_0)}\right)+ \int_{t=0}^T\frac{\|g^{(i-1)}(t,x_t)\|^2 }{2\sigma^2(t)}dt  \right ] }_{\text{(A2) KL\,Regularization\,relative\,to\,a diffusion at iteration $i-1$}  }  \nonumber 
 \end{align*}
For simplicity, we fix $\nu^{(i)} =\nu_{\pre}$. For methods to optimize initial distributions, refer to \citet{uehara2024finetuning}.

Here, suppose that $f$ is parameterized by $\theta$ (e.g., neural networks). Then, we update $\theta$ using stochastic gradient descent. Consider iteration $i  \in [1,2,\cdots,]$. With parameter $\theta$ fixed in $ f(t,x_t;\theta)$, by simulating an SDE with
\begin{align*}\textstyle
   dx_t &=  f(t,x_t;\theta) dt + \sigma(t) dw_t, x_0\sim  \nu^{\pre},  \\ 
   dz_t &=  \frac{\|g^{(0)}(t,x_t; \theta )\|^2}{2\sigma^2(t)} dt,\quad    dZ_t =  \frac{\|g^{(i-1)}(t,x_t; \theta)\|^2}{2\sigma^2(t)} dt, 
\end{align*}
we obtain $n$ trajectories
\begin{align*}\textstyle
    \{x^{\langle k \rangle}_0,\cdots,x^{\langle k \rangle}_T\}_{k=1}^n,\, \{z^{\langle k \rangle}_0,\cdots,z^{\langle k \rangle}_T\}_{k=1}^n,\, \{Z^{\langle k \rangle}_0,\cdots,Z^{\langle k \rangle}_T\}_{k=1}^n. 
\end{align*}
It is possible to use any off-the-shelf discretization methods, such as the Euler–Maruyama method. For instance, starting from $x^{\langle k \rangle}_0\sim \nu$, a trajectory can be obtained as follows:
\begin{align*}\textstyle
    &x^{\langle k \rangle}_t =x^{\langle k \rangle}_{t-1} +  f(t-1,x^{\langle k\rangle}_{t-1};\theta )\Delta t + \sigma(t) (\Delta w_t),\,\quad  \Delta w_t\sim \Ncal(0, (\Delta t)^2),  \\ 
    &z^{\langle k \rangle}_t= z^{\langle k \rangle}_{t-1} + \frac{\|f(t-1,x^{\langle k \rangle}_{t-1}; \theta ) - f^{\pre} (t-1,x^{\langle k \rangle}_{t-1} )\|^2}{2\sigma^2(t-1)} \Delta t, \\
    & Z^{\langle k \rangle}_t= Z^{\langle k \rangle}_{t-1} + \frac{\|f(t-1,x^{\langle k \rangle}_{t-1}; \theta ) - f^{(i-1)} (t-1,x^{\langle k \rangle}_{t-1};\theta^{(i-1)} ) \|^2}{2\sigma^2(t-1)}\Delta t. 
\end{align*}
Finally, using automatic differentiation, we update $\theta$ as follows:
\begin{align*}
   \theta_{i+1}=\theta_{i} - \rho \nabla_{\theta}\left\{ \frac{1}{n}\sum_{k=1}^n L(x^{\langle k \rangle}_T)-\alpha z^{\langle k \rangle}_T - \beta_{i-1} Z^{\langle k \rangle}_T \right \} \bigg{|}_{\theta=\theta_i},  
\end{align*}
where $\rho $ is a learning rate. For the practical selection of the learning rate $\rho$, we use the Adam optimizer \citep{kingma2014adam} in this step.

Note when discretization steps are large, the above calculation might be too computationally intensive. Regarding tricks to mitigate it, refer to \citep{clark2023directly,prabhudesai2023aligning}.

\section{Proofs}\label{sec:proof}

\subsection{Preliminary}\label{ape:prerlim}
In this section, we want to show how Assumption~\ref{assum:calibrated} holds when using linear models. A similar argument still holds when using GPs \citep{valko2013finite,srinivas2009gaussian}. 

Suppose we have a linear model: 
\begin{align*}
   \forall x \in \Xcal_{\pre};\,y = \mu^{\top}\phi(x) + \epsilon 
\end{align*}
where $\|\mu\|_2,\|\phi(x)\|_2 \leq B$, and $\epsilon$ is a $\sigma$-sub-Gaussian noise. Then, the ridge estimator is defined as 
\begin{align*}
   \hat \mu:= \Sigma^{-1}_i \sum_{j=1}^i \phi(x^{(j)}) y^{(j)},\quad \Sigma_i:=\sum_{j=1}^i \phi(x^{(j)})\phi^{\top}(x^{(j)}) + \lambda I. 
\end{align*}
With probability $1-\delta$, we can show that 
\begin{align*}
    \forall x\in \Xcal_{\pre};\,\langle \phi(x), \hat \mu - \mu\rangle\leq C_1(\delta) \|\phi(x)\|_{\Sigma_i}
\end{align*}
where 
\begin{align*}
    C_1(\delta):=  \left \{B\sqrt{\lambda} +\sqrt{ \sigma^2\left(\log (1/\delta^2) + d \log \left( 1 + \frac{K B^2}{d\lambda} \right )\right)} \right \}. 
\end{align*}
This is proved as follows. First, by some algebra, we have 
\begin{align*}
    \hat \mu - \mu = \lambda \Sigma^{-1}_i  \mu + \Sigma^{-1}_i\sum_{j=1}^i  \phi(x^{(j)})\epsilon_j. 
\end{align*}
Hence, with probability $1-\delta$, we can show that 
\begin{align*}
 \hat r(x)-r(x)   &= \langle \phi(x), \hat \mu - \mu\rangle\\
    &=\left \langle \phi(x), \lambda \Sigma^{-1}_i  \mu + \Sigma^{-1}_i\sum_j \phi(x^{(j)})\epsilon_j \right \rangle \\ 
    &\leq \|\phi(x)\|_{\Sigma_i}\|\lambda \mu\|_{\Sigma^{-1}_i } +\|\phi(x)\|_{\Sigma_i} \| \sum_j \phi(x^{(j)})\epsilon_j \|_{\Sigma^{-1}_i } \tag{CS inequality} \\
    &\leq   \|\phi(x)\|_{\Sigma_i}\left \{B\sqrt{\lambda} + \sqrt{ \sigma^2\left(\log (1/\delta^2) + d \log \left( 1 + \frac{iB^2}{d\lambda} \right )\right) }  \right \}. \tag{Use  Proof of Prop 6.7 and Lemma A.9 in \citet{agarwal2019reinforcement}}
\end{align*}
Hence, we can set 
\begin{align*}
    C_1(\delta) = B\sqrt{\lambda} + \sqrt{ \sigma^2\left(\log (1/\delta^2) + d \log \left( 1 + \frac{KB^2}{d\lambda} \right )\right)} . 
\end{align*}

\subsection{Proof of Theorem~\ref{thm:key}}

To simplify the notation, we let $f^{(i)}(x) =  \hat r^{(i)}(x)+ \hat g^{(i)}(x)$. 

We first note that 
\begin{align*}
      \mathrm{Term(A1)} &=\KL(\PP^{f,\nu}(\cdot)| \PP^{f^{\pre},\nu^{\pre}} (\cdot)). 
\end{align*}
Letting $\PP^{f,\nu}_{\cdot|0}$ be the distribution conditioned on an initial state $x_0$ (hence, $\nu$ does not matter), this is because 
\begin{align*}
     \KL(\PP^{f,\nu}(\cdot) \| \PP^{f^{\pre},\nu^{\pre}} (\cdot)) & = \KL(\nu \|\nu^{\pre}  ) + \EE_{x_0 \sim  \nu}[\KL(\PP^{f,\nu}_{\cdot|0}(\cdot |x_0) \| \PP^{f^{\pre},\nu^{\pre}}_{\cdot|0}(\cdot|x_0))]  \\
    &= \KL(\nu \|\nu^{\pre}  ) + \EE_{x_0 \sim  \nu}\left [\EE_{ \PP^{f,\nu}_{\cdot|0}(\cdot|x_0) }\left [\log\left(\frac{d\PP^{f,\nu}_{\cdot|0}(\cdot|x_0)}{d\PP^{f^{\pre},\nu^{\pre}}_{\cdot|0}(\cdot|x_0))} \right)\right]  \right] \\
    &= \KL(\nu \|\nu^{\pre}  ) + \EE_{\PP^{f,\nu}}\left[\int_{t=0}^T \frac{\|(f-f^{\pre})(t,x_t)\|^2_2}{2\sigma^2(t)} dt \right]. 
\end{align*}
Similarly, we have 
\begin{align*}
      \mathrm{Term(A2)} &=\KL(\PP^{f,\nu}(\cdot)| \PP^{f^{(i)},\nu^{(i)}} (\cdot)). 
\end{align*}
Therefore, the objective function in \eqref{eq:key} becomes
\begin{align*}
    \EE_{\PP^{f,\nu}}[r(x_T)] -\alpha \KL(\PP^{f,\nu}(\cdot) \| \PP^{f^{\pre},\nu^{\pre}} (\cdot))- \beta \KL(\PP^{f,\nu}(\cdot) \| \PP^{f^{(i-1)},\nu^{(i-1)}} (\cdot)).  
\end{align*}

Now, we further aim to modify the above objective function in \eqref{eq:key}. Here, letting $\PP^{f,\nu}_{\cdot|T}(\cdot|x_T)$ be the conditional distribution over $\Ccal$ conditioning on a state $x_T$,  we have 
\begin{align*}
     \KL(\PP^{f,\nu}(\cdot) \| \PP^{f^{\pre},\nu^{\pre}} (\cdot))  & = \int \log \left(\frac{d\PP^{f,\nu}(\cdot)}{d \PP^{f^{\pre},\nu^{\pre}} (\cdot)} \right)d\PP^{f,\nu}(\cdot) \\ 
     &= \int \log \left\{ \left(\frac{d\PP^{f,\nu}_T(x_T)}{d \PP^{f^{\pre},\nu^{\pre}}_T(x_T)} \right) + \left(\frac{d\PP^{f,\nu}_{\cdot|T}(\tau |x_T)}{d \PP^{f^{\pre},\nu^{\pre}}_{\cdot |T} (\tau|x_T)} \right) \right \} d\PP^{f,\nu}(\tau) \\
     &= \KL( \PP^{f,\nu}_T \| \PP^{f^{\pre},\nu^{\pre}}_T) + \EE_{x_T\sim \PP^{f,\nu}_T} [\KL(\PP^{f,\nu}_{\cdot|T}(\tau |x_T)  \|\PP^{f^{\pre},\nu^{\pre}}_{\cdot |T} (\tau|x_T) )  ].
\end{align*}
Hence, the objective function in \eqref{eq:key} becomes
\begin{align}\label{eq:maximize}
    &\underbrace{\EE_{x_T \sim \PP^{f,\nu}_T}\left [r(x_T)\right] -\alpha \KL( \PP^{f,\nu}_T \| \PP^{f^{\pre},\nu^{\pre}}_T) - \beta \KL( \PP^{f,\nu}_T \| \PP^{f^{(i-1)},\nu^{(i-1)}}_T ) }_{\text{(i)} }   \\ 
    &\underbrace{-\alpha \EE_{x_T \sim \PP^{f,\nu}_T }[\KL(\PP^{f,\nu}_{\cdot|T}(\cdot|x_T)\|\PP^{f^{\pre},\nu^{\pre}}_{\cdot|T}(\cdot|x_T))]}_{\text{(ii)}}\underbrace{-\beta \EE_{x_T \sim \PP^{f,\nu}_T }[\KL(\PP^{f,\nu}_{\cdot|T}(\cdot|x_T)\|\PP^{f^{(i-1)},\nu^{(i-1)}}_{\cdot|T}(\cdot|x_T))]}_{\text{(iii)}}.  \nonumber 
\end{align}

From now on, we use the induction method. Suppose we have 
\begin{align}\label{eq:induction}
        \PP^{f^{(i-1)},\nu^{(i-1)}}_{\cdot|T}(\cdot|x_T)= \PP^{f^{\pre},\nu^{\pre}}_{\cdot|T}(\cdot|x_T). 
\end{align}
Indeed, this holds when $i=1$. Assuming the above holds at $i-1$, the objective function in \eqref{eq:key} becomes
\begin{align}\label{eq:maximize2}
    &\underbrace{\EE_{x_T \sim \PP^{f,\nu}_T}\left [r(x_T)\right] -\alpha \KL( \PP^{f,\nu}_T \| \PP^{f^{\pre},\nu^{\pre}}_T) - \beta \KL( \PP^{f,\nu}_T \| \PP^{f^{(i-1)},\nu^{(i-1)}}_T ) }_{\text{(i)} }\\
    &\underbrace{-\alpha \EE_{x_T \sim \PP^{f,\nu}_T }[\KL(\PP^{f,\nu}_{\cdot|T}(\cdot|x_T)\|\PP^{f^{\pre},\nu^{\pre}}_{\cdot|T}(\cdot|x_T))]}_{\text{(ii)}}\underbrace{-\beta \EE_{x_T \sim \PP^{f,\nu}_T }[\KL(\PP^{f,\nu}_{\cdot|T}(\cdot|x_T)\|\PP^{f^{\pre},\nu^{\pre}}_{\cdot|T}(\cdot|x_T))]}_{\text{(iii)}}.  \nonumber 
\end{align}
By maximizing each term  $\text{(i)},\text{(ii)},\text{(iii)}$ in \eqref{eq:maximize2} over $\PP^{f,\nu}(\cdot,\diamond) = \PP^{f,\nu}_T(\diamond)\times \PP^{f,\nu}_{\cdot|T}(\cdot|\diamond)  $, we get 
\begin{align}
     \PP^{f^{(i)},\nu^{(i)}}_T(\diamond)  &\propto   \exp\left(\frac{\hat r^{(i)}(\diamond)+\hat g^{(i)}(\diamond)}{\alpha +\beta}\right)\{p^{(i-1)}(\diamond)\}^{\frac{\beta}{\alpha +\beta}}\{p_{\pre}(\diamond)\}^{\frac{\alpha}{\alpha +\beta}}, \nonumber  \\
     \PP^{f^{(i)},\nu^{(i)}}_{\cdot|T}(\cdot|x_T) &= \PP^{f^{\pre},\nu^{\pre}}_{\cdot|T}(\cdot|x_T).    \label{eq:bridge}
\end{align} 
Hence, the induction in \eqref{eq:induction} holds, and the statement is concluded.   

\begin{remark}
Note from the above, we can conclude that the whole distribution on $\Ccal$ is 
\begin{align*}
   \frac{1}{C} \exp\left(\frac{\hat r^{(i)}(x_T)+\hat g^{(i)}(x_T)}{\alpha +\beta}\right)\{p^{(i-1)}(\tau)\}^{\frac{\beta}{\alpha +\beta}}\{p_{\pre}(\tau)\}^{\frac{\alpha}{\alpha +\beta}}. 
\end{align*}    
\end{remark}

\begin{remark}
Some readers might wonder in the part we optimize over $\PP_{f,\nu}$ rather than $f,\nu$ in the final step. Indeed, this step would go through when we use non-Markovian drifts for $f$. While we use Markovian drifts, this is still fine because the optimal drift must be Markovian even when we optimize over non-Markovian drifts. We have chosen to present this intuitive proof first in order to more clearly convey our message regarding the bridge-preserving property \eqref{eq:bridge}. We will formalize it in \pref{thm:another_form} in  Section \ref{sec:detailed}.
\end{remark}

\subsection{Proof of Theorem~\ref{thm:regret}}

In the following, we condition on an event where
\begin{align*}
   \forall i \in [K];\,x \in \Xcal_{\pre}\,: |\hat r^{(i)}(x)-r(x)| \leq \hat g^{(i)}(x). 
\end{align*}
Then, we define 
\begin{align*}
    \hat f^{(i)}(x):=\hat r^{(i)}(x) + \hat g^{(i)}(x). 
\end{align*}
Recall that we have 
\begin{align}\label{eq:first_propety}
 \forall x\in \Xcal;  r(x) \leq \hat r^{(i)}(x) + \hat g^{(i)}(x)= \hat f^{(i)}(x). 
\end{align}
because when $x \in \Xcal_{\rho}$, it is from the assumption and when $x \notin \Xcal_{\rho}$, $r(x)$ takes $0$. Furthermore, 
we have 
\begin{align}\label{eq:second_propety}
  \forall x\in \Xcal_{\pre};\,  \hat f^{(i)}(x)-r(x)\leq \hat g^{(i)}(x) + \hat r^{(i)}(x)-r(x)\leq 2\hat g^{(i)}(x). 
\end{align}

With the above preparation, from now on, we aim to show the regret guarantee. First, by some algebra, we have 
\begin{align*}
  & 1/K \sum_{i=1}^K  \EE_{x \sim \pi}[r(x)]  - \EE_{x \sim p^{(i)}}[r(x)]   \\
  & = 1/K \sum_{i=1}^K  \EE_{x \sim \pi}[r(x)] - \EE_{x \sim \pi}[\hat f^{(i)}(x)] + \EE_{x \sim \pi}[\hat f^{(i)}(x)]-\EE_{x \sim p^{(i)}}[\hat f^{(i)}(x)]+\EE_{x \sim p^{(i)}}[\hat f^{(i)}(x)]- \EE_{x \sim p^{(i)} }[r(x)] \\
  &\leq  1/K \sum_{i=1}^K  \EE_{x \sim \pi}[\hat f^{(i)}(x)]-\EE_{x \sim p^{(i)}}[\hat f^{(i)}(x)]+\EE_{x \sim p^{(i)}}[\hat f^{(i)}(x)]- \EE_{x \sim p^{(i)} }[r(x)]   \tag{Optimism \eqref{eq:first_propety}} \\
  & \leq  1/K \sum_{i=1}^K { \EE_{x \sim \pi}[\hat f^{(i)}(x)]-\EE_{x \sim p^{(i)}}[\hat f^{(i)}(x)]}_{}+2 \EE_{x \sim p^{(i)}}[\hat g^{(i)}(x;\Dcal_{i})]. \tag{Use \pref{eq:second_propety} recalling the support of $p^{(i)}$ is in $\Xcal_{\pre}$ }
\end{align*}
Therefore, we have 
\begin{align*}
    & J_{\alpha}(\pi)-\frac{1}{K}\sum_i J_{\alpha}(p^{(i)} ) \\
    &=  \frac{1}{K} \sum_{i=1}^K \underbrace{ \EE_{x \sim \pi}[\hat f^{(i)}(x)]-\EE_{x \sim p^{(i)}}[\hat f^{(i)}(x)]-\alpha \KL(\pi\|\rho_{\pre}) + \alpha \KL( p^{(i)}\|\rho_{\pre})}_{\text{(To)}}+2 \EE_{x \sim p^{(i)}}[\hat g^{(i)}(x;\Dcal_{i})].
\end{align*}
Now, we analyze the term (To): 
\begin{align*}
   & \sum_{i=1}^K {\EE_{x \sim \pi}[\hat f^{(i)}(x)]-\EE_{x \sim p^{(i)}}[\hat f^{(i)}(x)]}-\alpha \KL(\pi\|\rho_{\pre}) + \alpha \KL( p^{(i)}\|\rho_{\pre}) \\ 
   &=\sum_{i=1}^K \beta_{i-1}\EE_{x\sim \pi}[\log(\pi(x)/p^{(i-1)}(x))]-(\beta_{i-1}+\alpha)\EE_{x\sim \pi}[\log(\pi(x)/p^{(i)}(x))] -\beta_{i-1}\EE_{x\sim p^{(i)}}[\log(p^{(i)}(x)/p^{(i-1)}(x))]\\
   &\leq \sum_{i=1}^K \beta_{i-1}\EE_{x\sim \pi}[\log(\pi(x)/p^{(i-1)}(x))]-(\beta_{i-1}+\alpha)\EE_{x\sim \pi}[\log(\pi(x)/p^{(i)}(x))]\\
   &\leq \beta_0 \EE_{x\sim \pi}[\log(\pi(x)/p^{(0)}(x))] \\ 
   &= 0. \tag{Recall we set $\beta_{i-1}=(i-1)\alpha$}
\end{align*}
Here, from the second line to the third line, we use 
\begin{align*}
     & {\EE_{x \sim \pi}[\hat f^{(i)}(x)]-\alpha \KL(\pi\|\rho_{\pre}) -\beta_{i-1}\KL(\pi \| p^{(i-1)})-\EE_{x \sim p^{(i)}}[\hat f^{(i)}(x)]} + \alpha \KL( p^{(i)}\|\rho_{\pre})  + \beta_{i-1}\KL(p^{(i)}\|p^{(i-1)} )  \\
    &=  \EE_{x \sim \pi}[\hat f^{(i)}(x)]-\alpha \KL(\pi\|\rho_{\pre}) -\beta_{i-1}\KL(\pi \| p^{(i-1)})-(\alpha + \beta_{i-1}) \int \hat f^{(i)}(x)\{  \rho_{\pre}(x)\}^{\alpha/\gamma}  \{ p^{(i-1)}(x)\}^{\beta_{i-1}/\gamma}dx  \\
    & = (\beta_{i-1} + \alpha) \EE_{x\sim \pi}[\log(\pi(x)/p^{(i)}(x))]. 
\end{align*}

Combining results so far, the proof is concluded. 
Therefore, we have 
\begin{align*}
     J_{\alpha}(\pi)-\frac{1}{K}\sum_i J_{\alpha}(p^{(i)} ) \leq  \frac{2}{K}\sum_i 
 \EE_{x \sim p^{(i)}}[\hat g^{(i)}(x;\Dcal_{i})].
\end{align*}

\subsection{Proof of Lemma \ref{lem:linear}}
Refer to the proof of Proposition 6.7 in \citet{agarwal2019reinforcement} in detail. For completeness, we sketch the proof here.  We denote
\begin{align*}
    \Sigma_i := \lambda I + \sum_i \phi(x^{(i)})\phi(x^{(i)})^{\top},\,  \hat h^{(i)}(x) :=  \min\left (1,  \sqrt{\phi(\cdot)^{\top}(\Sigma_i+\lambda I)^{-1} \phi(\cdot)}\right)
\end{align*}
Here, condoning on the event where Assumption~\ref{assum:calibrated} holds, 
we have 
\begin{align*}
     \underbrace{\EE\left [ \frac{1}{K}\sum_i \EE_{x\sim p^{(i)} }[\hat g^{(i)}(x) ] \right]}_{\textit{Expected regret }} & \leq 1/\sqrt{K}\times  \sqrt{ \sum_i \EE[\EE_{x\sim p^{(i)} }[ \{\hat g^{(i)}(x)\}^2] ]  }  \tag{Jensen's inequality} \\
    & =    \frac{C_1(\delta)}{\sqrt{K}}\times  \sqrt{  \sum_i \EE[ \EE_{x\sim p^{(i)} }[ \log (1 +    \{\hat h^{(i)}(x) \}^2)]]   } \tag{ $x \leq 2\log(1+x)$ when $1>x>0$ }\\
   & =    \frac{C_1(\delta)}{\sqrt{K}}\times  \sqrt{  2\sum_i \EE\left [ \log \left(1 + \phi^{\top}(x^{(i)})(\Sigma_i +\lambda )^{-1} \phi(x^{(i)})  \right)\right]    }  \\
    & =    \frac{C_1(\delta)}{\sqrt{K}} \times \sqrt{2  \sum_i \EE[ \log(\det \Sigma_{i} /\det \Sigma_{i-1}  )   ]  } \tag{Lemma 6.10 in \citet{agarwal2019reinforcement}  } \\
    & =   \frac{C_1(\delta)}{\sqrt{K}} \times \sqrt{2  \EE[ \log(\det \Sigma_{K} /\det \Sigma_{0}  )   ]  } \tag{Telescoping sum } \\
    & \leq    \frac{C_1(\delta)}{\sqrt{K}} \times \sqrt{2 d \log \left(1 + \frac{KB^2}{d\lambda }\right)} \tag{Elliptical potential lemma} 
\end{align*}
By taking $\delta=1/K^2$ and recalling 
\begin{align*}
    C_1(\delta) = B\sqrt{\lambda} + \sqrt{ \sigma^2\left(\log (1/\delta^2) + d \log \left( 1 + \frac{KB^2}{d\lambda} \right )\right)}, 
\end{align*}
the statement is concluded.

\section{More Detailed Theoretical Properties}\label{sec:detailed}

In this section, we show more detailed theoretical properties of fine-tuned diffusion models. In this proof, to simplify the notation, we denote $\beta_{i-1}$ by $\beta$. 

As a first step, we present a more analytical form of the optimal drift term by invoking the HJB equation, and the optimal value function. Note that we define the optimal value function by 
\begin{align*}
   v^{\star}_t(x_t) =  \EE_{\tau \sim \PP^{(i)} }\left[\hat r^{(i)}(x_T)+\hat g^{(i)}(x_T)
 - \int_{s=t}^T\frac{\beta \|(f^{(i)}-f^{(i-1)})(s,x_s)\|^2_2 +\alpha  \|(f^{\pre}-f^{(i)})(s,x_s)\|^2_2 }{2  \sigma^2(s) } ds |x_t \right ] 
\end{align*}

\begin{theorem}[Optimal drift and value function]\label{thm:optimal}

The optimal drift satisfies 
    \begin{align}\label{eq:optimal_form}
    u^{\star}(t,x) = \sigma^2(t)\frac{\nabla_x v^{\star}_t(x)}{\alpha + \beta} + \frac{ \{ \alpha f^{\pre} + \beta f^{(i-1)}\}(t,x) }{\alpha + \beta} 
\end{align}
where 
\begin{align*}
    \exp\left( \frac{v^{\star}_t(x)}{\alpha + \beta} \right)=\EE_{\PP^{\bar f}} \left[\exp\left(\frac{(\hat r^{(i)}+\hat g^{(i)})(x_T)}{\alpha + \beta} - \alpha \beta \int_{s=t}^T \frac{\|(f^{(i-1)}-f^{\pre})(s,x_s) \|^2}{2\sigma^2(s) } ds  \right) |x_t=x\right] 
\end{align*}  
and the measure $\PP^{\bar f}$ is induced by the following SDE: 
\begin{align*}
     d x_t = \bar f(t,x_t)dt  + \sigma(t)dw_t,\quad \bar f \coloneqq  \frac{\beta f^{(i-1)} + \alpha f^{\pre} }{\alpha + \beta},\quad  \bar \nu \coloneqq \frac{\beta \nu^{(i-1)}+ \alpha \nu^{\pre} }{\alpha + \beta}. 
\end{align*}
\end{theorem}

Using the above characterization, we can calculate the distribution induced by the optimal drift and initial distribution. 

\begin{theorem}[Formal statement of \pref{thm:key}]
\label{thm:another_form}
Let $\PP^{(i)}(\cdot)$ be a distribution over trajectories on $\Ccal$ induced by the diffusion model governed by $f^{(i)}, \nu^{(i)}$. Similarly, we define the conditional distribution over $\Ccal$ condoning on the terminal state $x_T$ by $\PP^{(i)}_{\cdot|T}(\cdot|x_T)$. Then, the following holds:  
\begin{align*}
  \PP^{(i)}(\tau)  &=   \exp\left(\frac{\hat r^{(i)}(x_T)+\hat g^{(i)}(x_T)}{\alpha +\beta}\right)\{\PP^{(i-1)}(\tau)\}^{\frac{\beta}{\alpha +\beta}}\{\PP_{\pre}(\tau)\}^{\frac{\alpha}{\alpha +\beta}},\\ 
  \PP^{(i)}_{\cdot|T}(\tau |x_T) &= \PP^{\pre}_{\cdot|T}(\tau |x_T), \\
  p^{(i)}(x_T) & \propto \exp\left(\frac{\hat r^{(i)}(x_T)+\hat g^{(i)}(x_T)}{\alpha +\beta}\right)\{p^{(i-1)}(x_T)\}^{\frac{\beta}{\alpha +\beta}}\{p_{\pre}(x_T)\}^{\frac{\alpha}{\alpha +\beta}}. 
\end{align*} 
\end{theorem}

\subsection{Proof of Theorem~\ref{thm:optimal}}

From the Hamilton–Jacobi–Bellman (HJB) equation, we have 
 \begin{align*}
     &\max_{u} \left \{\frac{\sigma^2(t)}{2}\sum_{i,j} \frac{d^2 v^{\star}_t(x)}{d x^{[i]}d x^{[j]}} +u \cdot \nabla v^{\star}_t(x) +\frac{d v^{\star}_t(x)}{d t}  - \frac{\alpha \|(u -f^{\pre})(t,x_t)\|^2_2}{2\sigma^2(t)}-\frac{\beta \|(u -f^{(i-1)})(t,x_t)\|^2_2}{2\sigma^2(t)}  \right \} =0. 
 \end{align*}
Here, we analyze the following terms:  
 \begin{align*}
      (a) = u \cdot \left \{ \nabla v^{\star}_t(x) + \frac{\alpha f^{\pre}(t,x_t) + \beta f^{(i-1)}(t,x_t)}{\sigma^2(t) }  \right \} -\frac{\alpha +\beta}{2\sigma^2(t)}\|u\|^2_2 +\frac{d v^{\star}_t(x)}{d t}-\frac{\alpha \|f^{\pre}(t,x_t)\|^2_2 + \beta \|f^{(i-1)}(t,x_t)\|^2_2 }{2\sigma^2(t)}. 
 \end{align*}
This is equal to  
 \begin{align*}
     (a)  &= - \frac{\alpha +\beta}{2\sigma^2(t)}\left \| u -  \frac{\sigma^2(t) \nabla v^{\star}_t(x)}{\alpha + \beta } -  \frac{\alpha f^{\pre} + \beta f^{(i-1)}}{\alpha + \beta }   \right \|^2 + (b),   \\
     (b) &= \frac{d v^{\star}_t(x)}{d t}-\frac{\alpha \|f^{\pre}\|^2_2 + \beta \|f^{(i-1)}\|^2_2 }{2\sigma^2(t)} + \frac{\alpha +\beta}{2\sigma^2(t)} \left \| \frac{\sigma^2(t) \nabla v^{\star}_t(x)}{\alpha + \beta } + \frac{\alpha f^{\pre} + \beta f^{(i-1)} }{ \alpha + \beta } \right \|^2 \\ 
   &= \frac{d v^{\star}_t(x)}{d t} +\frac{\sigma^2(t) \|\nabla v^{\star}_t(x)\|^2_2 }{2(\alpha+\beta)}- \frac{\alpha \beta   }{2\sigma^2(t)(\alpha + \beta) } \|(f^{\pre} - f^{(i-1)})(t,x) \|^2  + \frac{\nabla v^{\star}_t(x)\{\alpha f^{\pre} + \beta f^{(i-1)}\} }{ (\alpha + \beta) }. 
 \end{align*}
 Therefore, the HJB equation is reduced to 
 \begin{align}\label{eq:hjb}
     &\frac{\sigma^2(t)}{2}\sum_{i,j} \frac{d^2 v^{\star}_t(x)}{d x^{[i]}d x^{[j]}} +\frac{d v^{\star}_t(x)}{d t} +\frac{\sigma^2(t) \|\nabla v^{\star}_t(x)\|^2_2 }{2(\alpha+\beta)} \\ 
     &- \frac{\alpha \beta   }{2\sigma^2(t)(\alpha + \beta) } \|(f^{\pre} - f^{(i-1)})(t,x) \|^2   + \frac{\nabla v^{\star}_t(x)\{\alpha f^{\pre}(t,x) + \beta f^{(i-1)}(t,x)\} }{ (\alpha + \beta) }=0, \nonumber 
 \end{align}
 and the maximizer is 
 \begin{align*}
         u^{\star}(t,x) = \sigma^2(t)\frac{\nabla_x v^{\star}_t(x)}{\alpha + \beta} + \frac{ \alpha f^{\pre}(t,x) + \beta f^{(i-1)}(t,x)}{\alpha + \beta}. 
 \end{align*}
Using the above and denoting $\gamma = \alpha +\beta$, we can further show 
  \begin{align*}
  & \frac{\sigma^2(t)}{2}\sum_{i,j} \frac{d^2 \exp(v^{\star}_t(x)/\gamma) }{d x^{[i]}d x^{[j]}} + \frac{d\exp(v^{\star}_t(x)/\gamma)}{d t}  - \frac{\alpha \beta \exp(v^{\star}_t(x)/\gamma)   }{2\sigma^2(t)(\alpha + \beta)^2 }\|(f^{\pre} - f^{(i-1)})(t,x) \|^2 \\
  &+ \frac{\nabla \exp(v^{\star}_t(x)/\gamma ) \{\alpha f^{\pre}(t,x) + \beta f^{(i-1)}(t,x)\} }{(\alpha + \beta) }\\
  & = \frac{1}{\alpha+\beta}\exp\left (\frac{v^{\star}_t(x)}{\alpha + \beta } \right)  \left \{  \frac{\sigma^2(t)}{2}\sum_{i,j} \frac{d^2 v^{\star}_t(x)}{d x^{[i]}d x^{[j]}} +\frac{d v^{\star}_t(x)}{d t} +\frac{\sigma^2(t) \|\nabla v^{\star}_t(x)\|^2_2 }{2(\alpha+\beta)}  \right.\\
  & \left. - \frac{\alpha \beta   }{2\sigma^2(t)(\alpha + \beta) } \|(f^{\pre} - f^{(i-1)})(t,x) \|^2   + \frac{\nabla v^{\star}_t(x)\{\alpha f^{\pre}(t,x) + \beta f^{(i-1)}(t,x)\} }{ (\alpha + \beta) }  \right\}  \tag{Use \eqref{eq:hjb}} \\
  & = 0.  
 \end{align*}
Finally, using the Feynman–Kac formula, we obtain the form of the soft optimal value function:  
\begin{align*}
    \exp\left (\frac{v^{\star}_t(x_t)}{\alpha + \beta }\right) &= \EE_{\PP^{\bar f}}\left [\exp\left (\frac{(\hat r^{(i)} +\hat g^{(i)})(x_T)}{\alpha + \beta} -  \int_{t=s}^T \frac{\alpha \beta\|f^{\pre}(s,x_s)-f^{(i-1)}(s,x_s)\|^2_2 }{2(\alpha + \beta) ^2 \sigma^2(s) }ds \right )|x_t \right ],\\
    dx_t  &=  \bar f(t,x_t) d t + \sigma(t) dw_t,\quad \bar f(t,x_t) = \frac{\{\alpha f^{\pre} + \beta f^{(i-1)}\} (t,x_t)}{(\alpha + \beta) },\quad \bar \nu = \frac{\alpha \nu_{\pre } + \beta \nu^{\pre} }{\alpha + \beta},
\end{align*}
where the initial condition is 
\begin{align*}
    (\hat r^{(i)}+\hat g^{(i)})(x) = v^{\star}_T(x). 
\end{align*}

\subsection{Proof of \pref{thm:another_form}} 

We use induction. Suppose that the statement holds at $i-1$. This is indeed proven at $i=1$ in  \citet[Theorem 1]{uehara2024finetuning}. 
So, in the following, we assume 
\begin{align}\label{eq:inductive}
  \PP^{(i-1)}_{\cdot|T}(\tau |x_T) &= \PP^{\pre}_{\cdot|T}(\tau |x_T). 
\end{align} 

Under the above inductive assumption $i-1$, we first aim to prove the following:  
\begin{align}\label{eq:final_form}
   \exp\left( \frac{v^{\star}_t(x_t)}{\alpha + \beta} \right)= \int \exp\left(\frac{(\hat r^{(i)}+\hat g^{(i)})(x_T)}{\alpha + \beta}\right)\left \{ \frac{d \PP^{f^{\pre}}(x_T|x_t)}{d\mu  }\right\}^{\frac{\alpha}{\alpha+\beta}} \left \{\frac{d \PP^{f^{(i-1)}}(x_T|x_t)}{d\mu }\right\}^{\frac{\beta}{\alpha+\beta}} d\mu 
\end{align}
where $\mu$ is the Lebsgue measure. 

\paragraph{Proof of \eqref{eq:final_form}.}

To prove it, we use the following:  
 \begin{align}\label{eq:useful_one}
 1 = \int  \left \{ \frac{d \PP^{f^{\pre}}(x_{[t,T]}|x_t,x_T)}{d \PP^{\bar f}(x_{[t,T]}|x_t,x_T)} \right\}^{\frac{\alpha}{\alpha+\beta}} \left \{\frac{d \PP^{f^{(i-1)}}(x_{[t,T]}|x_t,x_T)}{d \PP^{\bar f}(x_{[t,T]}|x_t,x_T)} \right\}^{\frac{\beta}{\alpha+\beta}} \times d \PP^{\bar f}(x_{[t,T]}|x_t,x_T). 
\end{align}
This is proved by 
\begin{align*}
& \int  \left \{ \frac{d \PP^{f^{\pre}}(x_{[t,T]}|x_t,x_T)}{d \PP^{\bar f}(x_{[t,T]}|x_t, x_T)} \right\}^{\frac{\alpha}{\alpha+\beta}} \left \{\frac{d \PP^{f^{(i-1)}}(x_{[t,T]}|x_t,x_T)}{d \PP^{\bar f}(x_{[t,T]}|x_t,x_T)} \right\}^{\frac{\beta}{\alpha+\beta}} \times d \PP^{\bar f}(x_{[t,T]}|x_t,x_T)  \\ 
& = \int  \left \{ \frac{d \PP^{f^{\pre}}(x_{[t,T]}|x_t,x_T)}{d \PP^{\bar f}(x_{[t,T]}|x_t,x_T)} \right\}\times d \PP^{\bar f}(x_{[t,T]}|x_t,x_T)=1.  \tag{Use $    \frac{d \PP^{f^{\pre}}(x_{[t,T]}|x_t,x_T)}{d \PP^{f^{(i-1)}}(x_{[t,T]}|x_t,x_T)}=1 $ from the inductive assumption \eqref{eq:inductive}.  }
 \end{align*}
Then, using \pref{eq:useful_one}, this is proved as follows: 
{\small 
\begin{align*}
    (c)& :=    \int \exp\left(\frac{(\hat r^{(i)}+\hat g^{(i)})(x_T)}{\alpha + \beta}\right)\left \{ \frac{d \PP^{f^{\pre}}(x_T|x_t)}{d\mu  }\right\}^{\frac{\alpha}{\alpha+\beta}} \left \{\frac{d \PP^{f^{(i-1)}}(x_T|x_t)}{d\mu }\right\}^{\frac{\beta}{\alpha+\beta}} d\mu \\
     &= \int \exp\left(\frac{(\hat r^{(i)}+\hat g^{(i)})(x_T)}{\alpha + \beta}\right)\left \{ \frac{d \PP^{f^{\pre}}(x_T|x_t)}{d \PP^{\bar f}(x_T|x_t)} \right\}^{\frac{\alpha}{\alpha+\beta}} \left \{\frac{d \PP^{f^{(i-1)}}(x_T|x_t)}{d \PP^{\bar f}(x_T|x_t)} \right\}^{\frac{\beta}{\alpha+\beta}} d \PP^{\bar f}(x_T|x_t) \tag{Importance sampling} \\
     &= \int \exp\left(\frac{(\hat r^{(i)}+\hat g^{(i)})(x_T)}{\alpha + \beta}\right)\left \{ \frac{d \PP^{f^{\pre}}(x_{[t,T]}|x_t)}{d \PP^{\bar f}(x_{[t,T]}|x_t)} \right\}^{\frac{\alpha}{\alpha+\beta}} \left \{\frac{d \PP^{f^{(i-1)}}(x_{[t,T]}|x_t)}{d \PP^{\bar f}(x_{[t,T]}|x_t)} \right\}^{\frac{\beta}{\alpha+\beta}} \{d \PP^{\bar f}(x_T|x_t)\times d \PP^{\bar f}(x_{[t,T]}|x_t,x_T)\}. 
 \end{align*}
 } 
Here, from the second line to the third line, we use \eqref{eq:useful_one}, noticing
{\small 
\begin{align*}
    \frac{d \PP^{f^{\pre}}(x_{[t,T]}|x_t)}{d \PP^{\bar f}(x_{[t,T]}|x_t)} = \frac{d \{ \PP^{f^{\pre}}(x_{[t,T]}|x_t,x_T)\PP^{f^{\pre}}(x_{T}|x_t)\}  }{d \{ \PP^{\bar f}(x_{[t,T]}|x_t,x_T) \PP^{\bar f}(x_{T}|x_t) \} },\, \frac{d \PP^{f^{(i-1)}}(x_{[t,T]}|x_t)}{d \PP^{\bar f}(x_{[t,T]}|x_t)} = \frac{d \{ \PP^{f^{(i-1)}}(x_{[t,T]}|x_t,x_T)\PP^{f^{(i-1)}}(x_{T}|x_t)\}  }{d \{ \PP^{\bar f}(x_{[t,T]}|x_t,x_T) \PP^{\bar f}(x_{T}|x_t) \} }.
\end{align*}
} 
Finally, we have 
\begin{align*}
   (c)=  & \int \exp\left(\frac{(\hat r^{(i)}+\hat g^{(i)})(x_T)}{\alpha + \beta}\right)\left \{ \frac{d \PP^{f^{\pre}}(x_{[t,T]}|x_t)}{d \PP^{\bar f}(x_{[t,T]}|x_t)} \right\}^{\frac{\alpha}{\alpha+\beta}}   \left \{\frac{d \PP^{f^{(i-1)}}(x_{[t,T]}|x_t)}{d \PP^{\bar f}(x_{[t,T]}|x_t)} \right\}^{\frac{\beta}{\alpha+\beta}} d\PP^{\bar f}(x_{[t,T]}|x_t)   \\
     = & \int \exp\left(\frac{(\hat r^{(i)}+\hat g^{(i)})(x_T)}{\alpha + \beta}\right) \exp\left (- \int_{s=t}^T \frac{\alpha \beta\|(f^{\pre}(s,x_s)-f^{(i-1)})(s,x_s)\|^2_2 }{2(\alpha + \beta)^2 \sigma^2(s) }ds \right) d \PP^{\bar f}(x_{[t,T]}|x_t)\\
     = & \EE_{\PP^{\bar f}} \left[\exp\left(\frac{(\hat r^{(i)}+\hat g^{(i)})(x_T)}{\alpha + \beta}  - \alpha \beta \int_{s=t}^T \frac{\|(f^{(i-1)}-f^{\pre})(s,x_s) \|^2}{2\sigma^2(s) } ds  \right) |x_t=x\right]. 
\end{align*}
From the first line to the second line, we use the Girsanov theorem: 
{\small 
\begin{align*}
    \left \{ \frac{d \PP^{f^{\pre}}(x_{(t,T]}|x_t)}{d \PP^{\bar f}(x_{(t,T]}|x_t)} \right\}^{\frac{\alpha}{\alpha+\beta}} & = \exp\left (\frac{\alpha}{\alpha+\beta} \left\{\beta \int_{s=t}^T \frac{(f^{\pre}-f^{(i-1)})(s,x_s) }{\sigma(s)}dw_s  - \int_{s=t}^T \frac{\alpha \beta\|(f^{\pre}-f^{(i-1)})(s,x_s)\|^2_2 }{2(\alpha + \beta)^2 \sigma^2(s) }ds \right\} \right), \\
     \left \{\frac{d \PP^{f^{(i-1)}}(x_{(t,T]}|x_t)}{d \PP^{\bar f}(x_{(t,T]}|x_t)} \right\}^{\frac{\beta}{\alpha+\beta}} &= \exp\left (\frac{\beta}{\alpha+\beta} \left\{\alpha \int_{s=t}^T \frac{(f^{(i-1)} - f^{\pre})(s,x_s) }{\sigma(s)}dw_s  - \int_{s=t}^T \frac{\alpha \beta\|(f^{\pre}-f^{(i-1)})(s,x_s)\|^2_2 }{2(\alpha + \beta)^2 \sigma^2(s) }ds \right\} \right). 
\end{align*}
} 
Hence, using \pref{eq:optimal_form} in \pref{thm:optimal}, we can conclude
\begin{align*}
       (c)= \exp\left(\frac{v^{\star}_t(x_t)}{\alpha + \beta} \right). 
\end{align*}

\paragraph{Main part of the proof.} 

Now, we aim to prove the optimal distribution over trajectory $(\Ccal)$ is 
\begin{align*}
\frac{1}{C}\exp\left(\frac{\hat r^{(i)}(x_T)+\hat g^{(i)}(x_T)}{\alpha +\beta}\right)\{\PP^{(i-1)}(\tau)\}^{\frac{\beta}{\alpha +\beta}}\{\PP^{\pre}(\tau)\}^{\frac{\alpha}{\alpha +\beta}}.     
\end{align*}
To achieve this, we first show that the conditional optimal distribution over $\Ccal$ at state $x_0$ (i.e., $\PP^{(i-1)}_{\cdot|0}(\tau|x_0)$) is 
\begin{align}\label{eq:valid}
    \frac{1}{\exp\left (\frac{v^{\star}_0(x_0) }{\alpha + \beta} \right) }\times \exp\left(\frac{\hat r^{(i)}(x_T)+\hat g^{(i)}(x_T)}{\alpha +\beta}\right)\{\PP^{(i-1)}(\tau|x_0)\}^{\frac{\beta}{\alpha +\beta}}\{\PP^{\pre}(\tau|x_0)\}^{\frac{\alpha}{\alpha +\beta}}. 
\end{align}
First, we need to check that this is a valid distribution over $\Ccal$. This is because using an inductive hypothesis: 
\begin{align*}
    \PP^{\pre}(\tau |x_T,x_0) = \PP^{(i-1)}(\tau |x_T,x_0), 
\end{align*}
which is clear from \eqref{eq:inductive}, the above can be decomposed into   
\begin{align*}
\underbrace{ \left[ \frac{1}{ \exp\left (\frac{v^{\star}_0(x_0) }{\alpha + \beta} \right)  }\times \exp\left(\frac{\hat r^{(i)}(x_T)+\hat g^{(i)}(x_T)}{\alpha +\beta}\right)\{\PP^{(i-1)}(x_T|x_0)\}^{\frac{\beta}{\alpha +\beta}}\{\PP^{\pre}(x_T|x_0 )\}^{\frac{\alpha}{\alpha +\beta}}\right]}_{\text{$(\alpha 1)$} } \times \underbrace{\PP^{\pre} (\tau|x_T,x_0)}_{\text{$(\alpha 2)$} }. 
\end{align*}
Here, note that both term $(\alpha 1)$, $(\alpha 2)$ are normalized. Especially, to check $(\alpha 1)$ is normalized, we can use Equation~\eqref{eq:final_form} at $t=0$.

Now, after checking that \eqref{eq:valid} is a valid distribution, we calculate the KL divergence. This is calculated as follows:  
\begin{align*}
    & \KL\left(\PP^{(i)}_{\cdot|0}(\tau|x_0)\| \frac{1}{\exp\left (\frac{v^{\star}_0(x_0) }{\alpha + \beta} \right)  \nu^{\pre}(x_0) }\times \exp\left(\frac{\hat r^{(i)}(x_T)+\hat g^{(i)}(x_T)}{\alpha +\beta}\right)\{\PP^{(i-1)}(\tau)\}^{\frac{\beta}{\alpha +\beta}}\{\PP^{\pre}(\tau)\}^{\frac{\alpha}{\alpha +\beta}}\right ) \\
    &= \KL\left(\PP^{(i)}_{\cdot|0}(\tau|x_0)\|  \{\PP^{(i-1)}_{\cdot|0}(\tau|x_0)\}^{\frac{\beta}{\alpha +\beta}}\{\PP^{\pre}(\tau|x_0)\}^{\frac{\alpha}{\alpha +\beta}}\right )
    + \EE_{\tau \sim \PP^{(i)} }\left[ \left(\frac{\hat r^{(i)}(x_T)+\hat g^{(i)}(x_T)}{\alpha +\beta}\right)  \right ] - \frac{v^{\star}_0(x_0)}{\alpha + \beta } \\ 
    &=\EE_{\tau \sim \PP^{(i)} }\left[\left(\frac{\hat r^{(i)}(x_T)+\hat g^{(i)}(x_T)}{\alpha +\beta}\right) 
 - \int_{s=0}^T\frac{\beta \|(f^{(i)}-f^{(i-1)})(s,x_s)\|^2_2 +\alpha  \|(f^{\pre}-f^{(i)})(s,x_s)\|^2_2 }{2(\alpha + \beta) \sigma^2(s) }ds   \right ] - \frac{v^{\star}_0(x_0)}{\alpha + \beta} \tag{Use Girsanov theorem} \\
 & =0 \tag{Use Definition of soft optimal value functions}
\end{align*}
Hence, we can now conclude \eqref{eq:valid}. 

Next, we consider an exact formulation of the optimal initial distribution. We just need to solve 
\begin{align*}
    \argmax_{\nu \in \Delta(\Xcal) }\int v^{\star}_0(x)\nu (x)dx -\beta \KL(\nu \|\nu^{(i-1)})- \alpha \KL(\nu \|\nu^{\pre}). 
\end{align*}
The closed-form solution is proportional to  
\begin{align}
\label{eq:optimal}
   \frac{1}{C} \exp\left (\frac{v^{\star}_0(x)}{\alpha + \beta}  \right) \{ \nu^{(i-1)}(x)\}^{ \frac{\alpha}{\alpha + \beta} } \nu^{\pre}(x)\}^{\frac{\alpha}{\alpha + \beta} }. 
\end{align}

Finally, by multiplying \eqref{eq:optimal} and \eqref{eq:valid},  we can conclude that the optimal distribution over $\Ccal$ is 
\begin{align*}
    \frac{1}{C}\exp\left(\frac{\hat r^{(i)}(x_T)+\hat g^{(i)}(x_T)}{\alpha +\beta}\right)\{\PP^{(i-1)}(\tau)\}^{\frac{\beta}{\alpha +\beta}}\{\PP^{\pre}(\tau)\}^{\frac{\alpha}{\alpha +\beta}}.  
\end{align*}

\section{Experiment Details}\label{ape:experiments}

\subsection{Implementation of Baselines}\label{subsec:baselines}

In this section, we describe more details of the baselines. 

\paragraph{Online PPO.}

Considering the discretized formulation of diffusion models  \citep{black2023training,fan2023dpok}, we use the following update rule:
\begin{align} \label{eq:natural}
  & \nabla_{\theta}\EE_{\Dcal}\sum_{t=1}^T \left [\min\left \{\tilde r_t(x_0,x_t)\frac{p(x_t|x_{t-1};\theta)}{p(x_t|x_{t-1};\theta_{\text{old}})}, \tilde r_t(x_0,x_t) \cdot \mathrm{Clip}\left (\frac{p(x_t|x_{t-1};\theta)}{p(x_t|x_{t-1};\theta_{\text{old}})}, 1-\epsilon,1+\epsilon\right ) \right\} \right],   \\
   & \tilde r_t(x_0,x_t) = - r(x_T) + \underbrace{\alpha \frac{\|u(t,x_t;\theta)\|^2 }{2 \sigma^2(t)}}_{\text{KL term}},\quad p(x_t|x_{t-1};\theta)=\Ncal(u(t,x_t;\theta)+f(t,x_t),\sigma(t)) \label{eq:natural_2}
\end{align}
Here, note a pre-trained diffusion model is denoted by $p(x_t|x_{t-1};\theta)=\Ncal({f(t,x_{t-1})},\sigma )$ and $\theta$ is a parameter to be optimized. 

Note that DPOK~\citep{fan2023dpok} uses the following update: 
\begin{align*} 
  & \nabla_{\theta}\EE_{\Dcal}\sum_{t=1}^T \left [\min\left \{ -r(x_0)\frac{p(x_t|x_{t-1};\theta)}{p(x_t|x_{t-1};\theta_{\text{old}})},  - r(x_0) \cdot \mathrm{Clip}\left (\frac{p(x_t|x_{t-1};\theta)}{p(x_t|x_{t-1};\theta_{\text{old}})}, 1-\epsilon,1+\epsilon\right )\right\} +  \underbrace{\alpha \frac{\|u(t,x_t;\theta)\|^2 }{2\sigma^2(t)}}_{\text{KL term} } \right]
\end{align*}
where the KL term is directly differentiated. We did not use the DPOK update rule because DDPO appears to outperform DPOK even without a KL penalty (\citet{black2023training}, Appendix C), so we implemented this baseline by modifying the DDPO codebase to include the added KL penalty term (Equation~\eqref{eq:natural_2}).

\paragraph{Guidance.}

We use the following implementation of guidance \citep{dhariwal2021diffusion}: 
\begin{itemize}
\item For each $t \in [0,T]$, we train a model: $p(y|x_t)$ where $x_t$ is a random variable induced by the pre-trained diffusion model. 
    \item We fix a guidance level $\gamma \in \RR_{>0}$, target value $y_{\condition} \in \RR$, and at 
    inference time (during each sampling step), we use the following score function 
\begin{align*}
    \nabla_x \log{p(x|y = y_{\condition})} = \nabla_x \log{p(x)} + \gamma \nabla_x \log{p(y=y_{\condition}|x)}. 
\end{align*}
\end{itemize}

A remaining question is how to model $p(y|x)$. In our case, for the biological example, we make a label depending on whether $x$ is top $10\%$ or not and train a binary classifier. In image experiments, we construct a Gaussian model: $p(y|x)=\mathcal{N}(y-\mu_\theta(x), \sigma^2)$ where $y$ is the reward label, $\mu_\theta$ is the reward model we need to train, and $\sigma$ is a fixed hyperparameter.

\subsection{Experiment in Molecules}\label{subsec:molecules}

We do an analogous experiment in Section~\ref{subsec:biology} to generate molecules with improved properties, specifically the Quantitative Estimate of Druglikeness (QED) score. It's important to note that this experiment is conducted on a simplified scale, as we utilize a trained oracle instead of a real black box oracle for evaluation.

This focuses on obtaining a module with favorable properties. Here, $x$ represents the molecule, and $y$ corresponds to the Quantitative Estimate of Druglikeness (QED) score. We employ a pre-trained diffusion model using the ZINC dataset \citep{irwin2005zinc}. In this case, we use a graph neural network-based diffusion \citep{jo2022score}. To quantify diversity, we employ the $(1-\text{Tanimoto coefficient})$ of the standard molecular fingerprints \citep{jo2022score,bajusz2015tanimoto}. We set $M_i = 2500$ and $K=4$ ($M=10000$). 

\begin{figure}[!h]
    \centering
    \includegraphics[width = 0.5\linewidth]{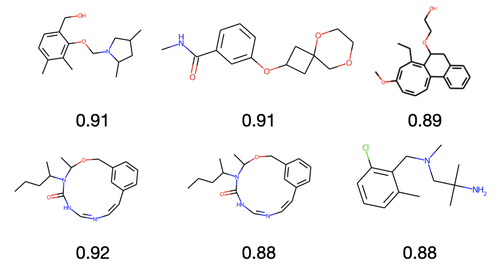}
    \caption{Examples of molecules that attain high QED generated by \alg.}
    \label{fig:mol}
\end{figure}

The results are reported in \pref{tab:ZINC} and \pref{fig:mol}. We observe similar trends in Section~\ref{subsec:biology}.

\begin{table}[!h]
    \centering
      \caption{Results for fine-tuning diffusion models for molecules using the ZINC dataset to optimize QED scores. \alg\, attains high rewards using a fixed budget of feedback.}
    \begin{tabular}{c c   c} \toprule 
         & Value $\uparrow$  &  Div $\uparrow$    \\  \midrule 
    \textbf{Non-adaptive} & $0.82\pm 0.02$  &  $0.85$   \\
    \textbf{Guidance} & $0.73\pm 0.02$    &   $\mathbf{0.88}$   \\ \hline 
     \textbf{Online PPO} &  $0.76 \pm 0.03$     &  $0.85$\\   
    \textbf{Greedy} &  $0.72\pm 0.00$   & $0.74$ \\  \hline
  \rowcolor{Gray}   \textbf{UCB (Ours)} & $0.86 \pm 0.01 $  &   $0.86$  \\
  \rowcolor{Gray}   \textbf{Bootstrap (Ours)}  & $\mathbf{0.88 \pm 0.01 }$   &  $0.83$\\ \bottomrule
    \end{tabular}
    \label{tab:ZINC}  
\end{table}

\subsection{Details for Protein Sequences and Molecules}

\subsubsection{Description of Data}

\paragraph{GFP.} The original dataset size is $56086$. Each data consists of an amino acid sequence with $8$-length. We represent each data as a one-hot encoding representation with dimension $237\times 20$. Here, we model the difference between the original and baseline sequences. We selected the top $33637$ samples following \citet{trabucco2022design} and trained diffusion models and oracles using this selected data.  

\paragraph{ZINC.} The ZINC dataset for molecules is a large and freely accessible collection of chemical compounds used in drug discovery and computational chemistry research \citep{irwin2005zinc}. The dataset contains a diverse and extensive collection of chemical structures, including small organic molecules.    

The dataset size we use is $249,455$. We use the entire dataset to acquire pre-trained diffusion models. For the numerical representation of each molecule, we adopt a graph representation denoted as $x \in \RR^{38 \times 47}$. This representation comprises two matrices: $x_1 \in \RR^{38 \times 38}$ and $x_2 \in \RR^{38 \times 9}$. A matrix $x_1$ serves as an adjacency matrix for the molecule, while $x_2$ encodes features for each node. With $38$ nodes, in total, each node is associated with a $9$-dimensional feature vector representing its correspondence to a specific element.

\subsubsection{Architecture of Neural Networks}

 Concerning diffusion models and fine-tuning for ZINC, we adopt the architecture presented in \citet{jo2022score}, specifically designed for graph representation. They have implemented this architecture by effectively stacking transformed and GFN components. Therefore, we will not delve into the details here.

Regarding diffusion models and fine-tuning for GFP, we add detail for protein sequences in Table~\ref{tab:diffusion_tfbind}, \ref{tab:oralce_TFBind}.

\begin{table}[!t]
    \centering
        \caption{Architecture of diffusion models for GFP }    \label{tab:diffusion_tfbind}
    \begin{tabular}{c c c c }
    \toprule 
     Layer    &  Input Dimension  & Output dimension  & Explanation \\ \hline 
      1   &    $1$ $(t)$ &   $256$ ($t'$)    & Get time feature  \\
      1   & $237 \times 20 $ $(x)$ &   $64$ ($x'$)     & Get positional encoder  (Denote $x'$)  \\ 
    2  &  $237\times 20 + 256 + 64$ $(x,t,x')$  &  $64$ $(\bar x)$  & Transformer Encoder  \\
    3 & $ 64$  $(\bar x)$ &   $237 \times 20  $  $(x)$ &  Linear 
     \\ \bottomrule  
     \end{tabular}
\end{table}

\begin{table}[!t]
    \centering
      \caption{Architecture of oracles for GFP}
    \label{tab:oralce_TFBind}
    \begin{tabular}{cccc} \toprule 
         &  Input dimension & Output dimension &  Explanation \\  \hline 
      1   & $237\times 20 $   &   $500$  & Linear \\ 
      1   &  $500$ &  $500$  & ReLU  \\ 
      2  &  $500$ &  $200$   & Linear \\
     2    &  $200$ &  $200$  & ReLU  \\ 
    3  &  $200$ &  $1$   & Linear \\
     3    &  $200$ &  $1$  & ReLU  \\
     4  & $1$ & $1$  & Sigmoid \\ \bottomrule 
    \end{tabular}
\end{table}

\subsubsection{Hyperparameters}

In all our implementations, we utilize A100 GPUs. For the fine-tuning of diffusion models, we employ the specific set of hyperparameters outlined in Table~\ref{tab:hyper}.

\begin{table}[!t]
    \centering
     \caption{Important hyperparameters for fine-tuning. For all methods, we use ADAM as an optimizer. }
    \label{tab:hyper}
    \begin{tabular}{c|c|cc}\toprule 
   Method  &   Type     &  GFP  
& ZINC   \\  \hline 
  \multirow{8}{*}{\alg}  &   Batch size  & $128$ & $32$ \\
  & KL parameter $\beta$  & $0.01$  &  $0.002$ \\ 
  & UCB parameter $C_1$ & $0.002$  &  $0.002$   \\  
   & Number of bootstrap heads & $3$ & $3$ \\  
   & Sampling to neural SDE &  \multicolumn{2}{c}{Euler~~~~Maruyama} \\ 
     & Step size (fine-tuning)  & $50$ & $25$  \\ 
     & Epochs (fine-tuning) & $100$ &  $100$ \\
     \hline 
 \multirow{3}{*}{\textbf{PPO}}   &  Batch size & $128$ & $128$ \\ 
 & $\epsilon$ & $0.1$ & $0.1$\\
  & Epochs & $100$ & $100$ \\  
  \hline  
 \textbf{Guidance}  & Guidance level  &  10 & 10 \\ \hline  
 \textbf{Pre-trained diffusion}  & Forward SDE  & \multicolumn{2}{c}{Variance preserving} \\
 & Sampling way &  \multicolumn{2}{c}{Euler~~~~Maruyama}\\  
  \bottomrule  
    \end{tabular}
\end{table}

\subsection{Details for Image Tasks}

 \paragraph{Prompts.} Since StableDiffusion is a text-to-image model, we need to specify prompts used for training and evaluation, respectively. To align with prior studies, our training process involves using prompts from a predefined list of 50 animals~\citep{black2023training,prabhudesai2023aligning}. In evaluation, we adopt the following animals: snail, hippopotamus, cheetah, crocodile, lobster, and octopus; all are not seen during training.

\paragraph{Techniques for saving CUDA memory.} We use the DDIM sampler~\citep{song2020denoising} with 50 steps for sampling. To manage memory constraints during back-propagation through the sampling process and the VAE decoder, we implemented two solutions from \citet{clark2023directly,prabhudesai2023aligning}: (1) Fine-tuning LoRA modules~\citep{hu2021lora} instead of the full diffusion weights, and (2) Using gradient checkpointing~\citep{gruslys2016memory, chen2016training}. We also applied randomized truncated back-propagation, limiting gradient back-propagation to a random number of steps denoted as $K$. In practice, $K$ is uniformly obtained from $(0,50)$ as~\citet{prabhudesai2023aligning}.

\begin{table}[!t]
    \centering
     \caption{Important hyperparameters for fine-tuning Aesthetic Scores. }
    \label{tab:hyper_params}
    \begin{tabular}{c|c|cc}\toprule 
  \multirow{9}{*}{\alg}  &   Batch size per GPU  & $2$\\
  & Samples per iteration  & $64$   \\
  & Samples per epoch  & $128$   \\ 
  & KL parameter $\beta$  & $1$   \\ 
  & UCB parameter $C_1$ & $0.01$  \\
  & UCB parameter $\lambda$ & $0.001$  \\  
   & Number of bootstrap heads & $4$  \\  
     & DDIM Steps  & $50$  \\ 
     & Guidance weight & $7.5$ \\
     \hline 
\multirow{6}{*}{ \textbf{PPO}}   &  Batch size per GPU & $4$ \\
 &  Samples per iteration & $128$ \\
 &  Samples per epoch & $256$ \\
 & KL parameter $\beta$ & 0.001\\
 & $\epsilon$ & 1e-4\\
  & Epochs & $60$ \\  
  \hline  
 \textbf{Guidance}  & Guidance level  &  400 \\ \hline  
\multirow{6}{*}{ \textbf{Optimization} } & Optimizer  & AdamW \\
 & Learning rate  & 1e-4 \\
 & $(\epsilon_1, \epsilon_2)$ &  $(0.9, 0.999)$\\ 
 & Weight decay & 0.1\\
 & Clip grad norm & 5\\
 & Truncated back-propagation step $K$ & $K \sim \text{Uniform}(0,50)$\\
  \bottomrule  
    \end{tabular}
\end{table}

\paragraph{Guidance.}

To approximate the classifier $p(y|x_t)$, we train a reward model $\mu_\theta(x_t,t)$ on the top of the OpenAI CLIP embeddings~\citep{radford2021clip}. The reward model is implemented as an MDP, which inputs the concatenation of sinusoidal time embeddings (for time $t$) and CLIP embeddings (for $x_t$). Our implementation is based on the public RCGDM~\citep{yuan2023reward} codebase\footnote{\url{https://github.com/Kaffaljidhmah2/RCGDM}}.

\paragraph{Online PPO.} We implement a PPO-based algorithm, specifically DPOK~\citep{fan2023dpok}, configured with $M=15360$ feedback interactions. The implementation is based on the public DDPO~\citep{black2023training} codebase\footnote{\url{https://github.com/kvablack/ddpo-pytorch}}.

\subsubsection{Hyperparameters}
In all image experiments, we use four A100 GPUs for fine-tuning StableDiffusion v1.5~\citep{Rombach_2022_CVPR}. Full training hyperparameters are listed in Table~\ref{tab:hyper_params}

\subsubsection{Additional Results}

\paragraph{Training Curves.}

\begin{figure}[!h]
\centering
\includegraphics[width = 0.8\linewidth]{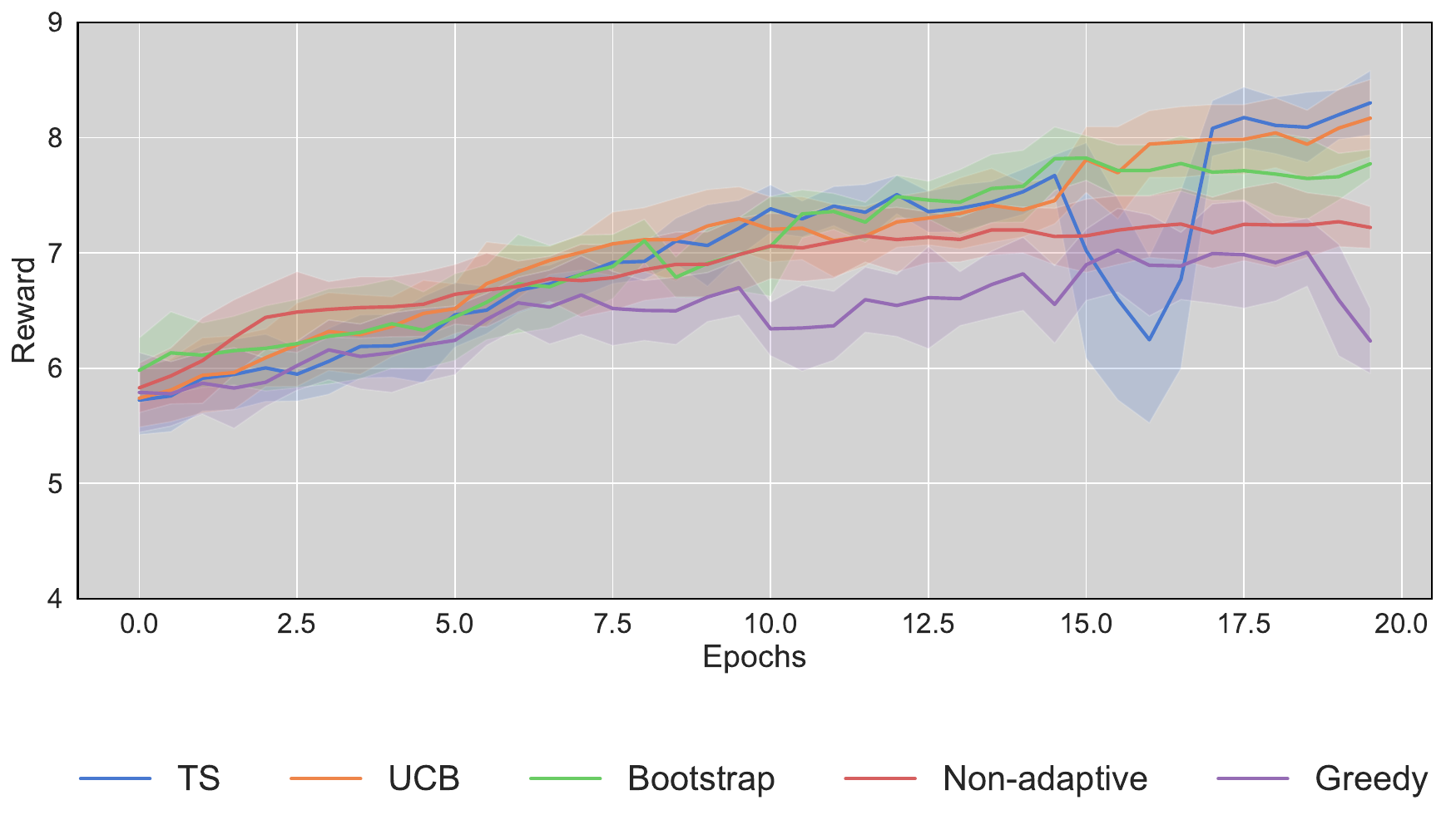}
\caption{Training curves of reward $r(x)$ for fine-tuning aesthetic scores.}
\label{fig:training}
\end{figure}

We plot the training curves in Figure~\ref{fig:training}. Recall we perform a four-shot online fine-tuning process. For each online iteration, the model is trained with $5$ epochs. In Table~\ref{tab:fig_table}, we report the final results after $20$ epochs. 

\begin{figure}[!t]
    \centering
\includegraphics[width=\linewidth]{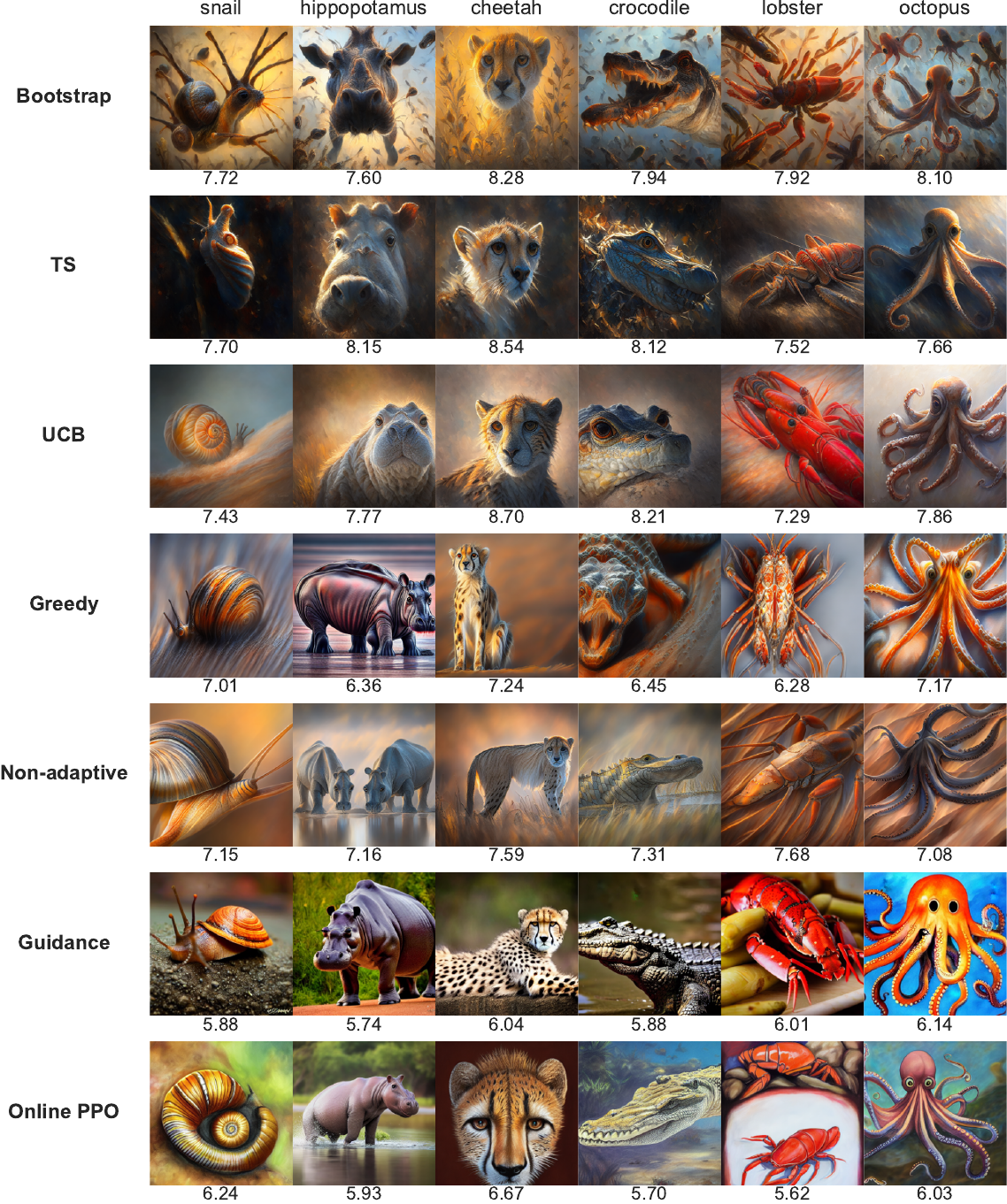}
    \caption{More images generated by $\alg$ and baselines. For all algorithms, fine-tuning is conducted with a total of $15600$ samples.}
    \label{fig:more-samples}
\end{figure}

\paragraph{Additional Generated Images.}
In Figure~\ref{fig:more-samples}, we provide more qualitative samples to illustrate the performance of $\alg$ in fine-tuning aesthetic quality. 

